%% file: 0_main.tex
\documentclass[11pt]{article}

\usepackage[final]{acl}

\usepackage{times}
\usepackage{latexsym}

\usepackage[T1]{fontenc}

\usepackage[utf8]{inputenc}

\usepackage{microtype}

\usepackage{inconsolata}

\usepackage{graphicx}

\usepackage{url}            
\usepackage{booktabs}       
\usepackage{amsfonts}       
\usepackage{nicefrac}       
\usepackage{xcolor}         
\usepackage{amsmath}
\usepackage{amssymb}
\usepackage{amsfonts}
\usepackage{nomencl}
\usepackage{glossaries}
\usepackage{xspace}
\usepackage{multirow}
\usepackage{pifont}
\usepackage{algorithm}
\usepackage{subcaption}
\usepackage{enumitem}
\usepackage{multicol}
\usepackage{caption}
\usepackage{array}
\usepackage{fp}
\usepackage{makecell}
\usepackage{cases}
\usepackage{color}
\usepackage{algpseudocode}
\usepackage{listings}
\usepackage{mathrsfs}
\usepackage{longtable}
\usepackage{colortbl}
\usepackage{bm}
\usepackage{makecell}
\usepackage{bbm}
\usepackage{arydshln}
\usepackage{booktabs}
\usepackage[most]{tcolorbox}
\usepackage{footnote}
\usepackage{needspace}   
\newcommand{\ignore}[1]{\iffalse#1\fi}

\newcommand{\eg}{\emph{e.g.,}\xspace}

\newcommand{\TaskEntry}[4]{%
  \begin{savenotes}%
    \par\addvspace{0.8em}%
    \noindent\paragraph{#1}%
    #2\par
    \addvspace{0.4em}%
    \Needspace{8\baselineskip}%
    \begin{center}
      \setlength{\fboxrule}{1pt}%
      \fbox{\includegraphics[width=0.95\linewidth]{#3}}%
      \captionof{figure}{Example image of task #1}%
    \end{center}
    \begin{tcolorbox}[
      enhanced,
      colback=gray!5!white,
      colframe=gray!55!black,
      width=0.98\linewidth,
      title={{Task prompt for #1}},
      boxsep=0.8mm,
      left=1mm,right=1mm,top=1mm,bottom=1mm
    ]
      #4
    \end{tcolorbox}
  \end{savenotes}%
  \par\addvspace{0.8em}%
}

\colorlet{boxframe}{gray!55!black}
\title{Beyond the Last Frame: Process-aware Evaluation for \\ Generative Video Reasoning}

\author{
\setcounter{footnote}{1}
	Yifan Li\textsuperscript{\rm{1},\rm{3},\footnotemark[1],\footnotemark[2]},
    Yukai Gu \textsuperscript{\rm{1},\rm{3},\footnotemark[1],\footnotemark[2]},
    Yingqian Min\textsuperscript{\rm{1},\rm{3}},
    Zikang Liu\textsuperscript{\rm{1},\rm{3}},
    Yifan Du\textsuperscript{\rm{1},\rm{3}},
	Kun Zhou\textsuperscript{\rm{2}},\\
    \textbf{Min Yang} \textsuperscript{\rm{4}},
	\textbf{Wayne Xin Zhao}\textsuperscript{\rm{1},\rm{3},\footnotemark[3]} \and \textbf{Minghui Qiu} \textsuperscript{\rm{4},\footnotemark[3]}
	\\
        \textsuperscript{1}Gaoling School of Artificial Intelligence, Renmin University of China \\
        \textsuperscript{2}School of Information, Renmin University of China \\
	\textsuperscript{3}Beijing Key Laboratory of Research on Large Models and Intelligent Governance \\
	\textsuperscript{4}ByteDance\\
	 \texttt{\{liyifan0925, batmanfly\}@gmail.com} \\
}

\begin{document}
\maketitle
\footnotetext[1]{Equal contribution.}
\footnotetext[2]{Work done during internship at ByteDance.}
\footnotetext[3]{Corresponding author.}
\begin{abstract}
Recent breakthroughs in video generation have demonstrated an emerging capability termed Chain-of-Frames (CoF) reasoning, where models resolve complex tasks through the generation of continuous frames. While these models show promise for Generative Video Reasoning (GVR), existing evaluation frameworks often rely on single-frame assessments, which can lead to outcome-hacking, where a model reaches a correct conclusion through an erroneous process. To address this, we propose a process-aware evaluation paradigm. We introduce VIPER, a comprehensive benchmark spanning 16 tasks across temporal, structural, symbolic, spatial, physics, and planning reasoning. Furthermore, we propose Process-outcome Consistency (POC$@r$), a new metric that utilizes VLM-as-Judge with a hierarchical rubric to evaluate both the validity of the intermediate steps and the final result. Our experiments reveal that state-of-the-art video models achieve POC$@$1.0 only about ~20\% and exhibit a significant outcome-hacking. We further explore the impact of test-time scaling and sampling robustness, highlighting a substantial gap between current video generation and true generalized visual reasoning. Our benchmark are released at \url{https://github.com/RUCAIBox/VIPER}.
\end{abstract}

\input{latex/1_intro}
\input{latex/2_empirical}

\input{latex/3_benchmark}
\input{latex/4_evaluation}

\input{latex/5_experiment}
\input{latex/6_related_work}
\input{latex/7_conclusion}

\bibliography{custom}

\appendix

\input{latex/8_appendix}

\end{document}

%% file: latex/1_intro.tex
\section{Introduction}

Recent breakthroughs in video generation have enabled models to produce highly realistic, consistent, and extended sequences of real-world scenes \cite{sora2, veo3, chen2025seedance}. While these advancements demonstrate immense potential for artistic creation and digital media, recent research~\cite{Noam2025pathways, wiedemer2025veo3-zeroshot} suggests that their impact extends far beyond content generation: video models are beginning to exhibit generalized visual reasoning capabilities. Specifically, by generating continuous sequences of frames, video models can depict the step-by-step resolution of tasks ranging from fundamental visual processing (\eg super-resolution) to high-level logical reasoning (\eg maze solving). This emerging capability is termed ``Chain-of-Frames'' (CoF) reasoning.

\begin{figure}[t]
    \centering
    \includegraphics[width=0.9\linewidth]{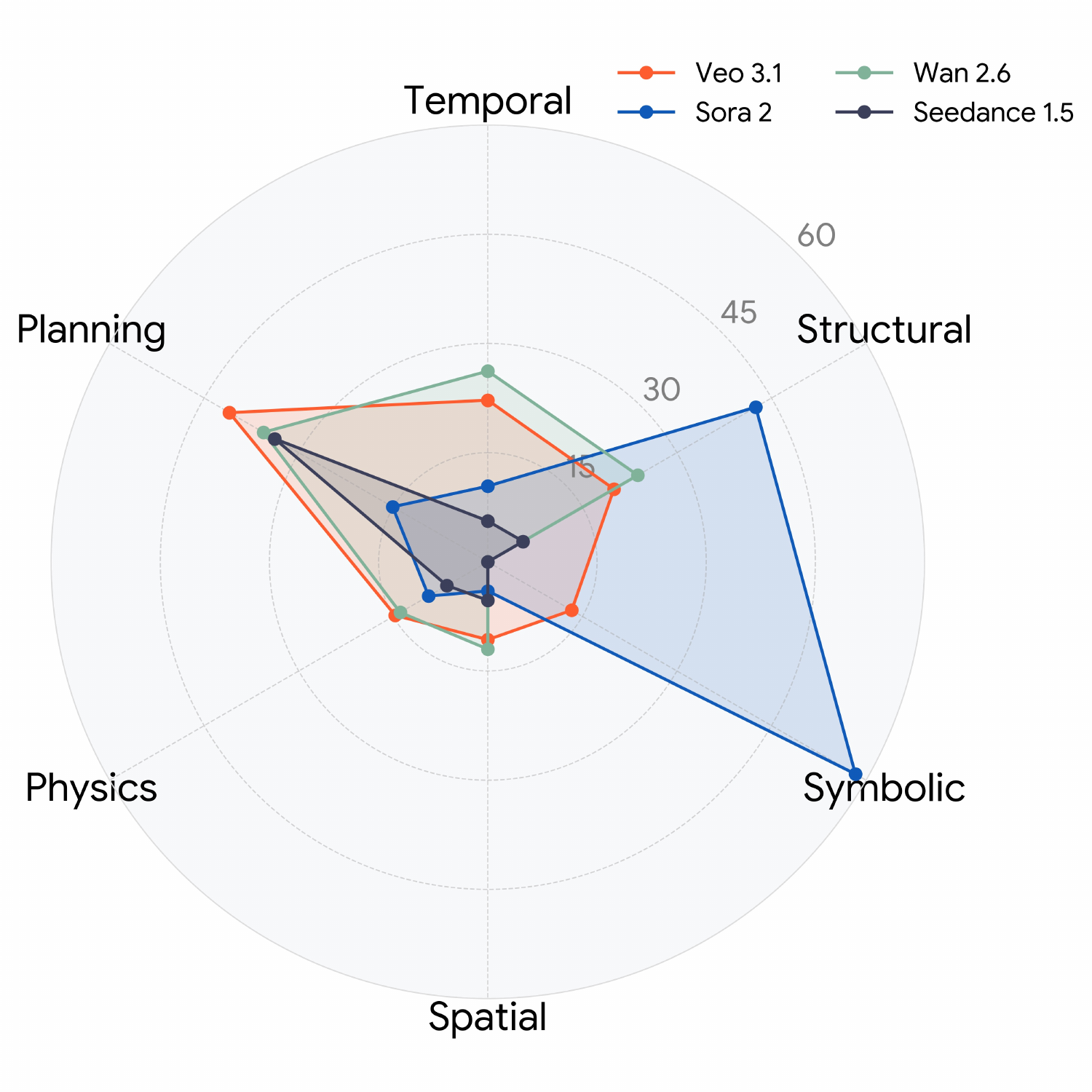}
    \caption{POC@1.0 performance overview of representative video models on VIPER across 6 domains.}
    \label{fig:radar}
\end{figure}

\begin{figure*}[t]
    \centering
    \includegraphics[width=\linewidth]{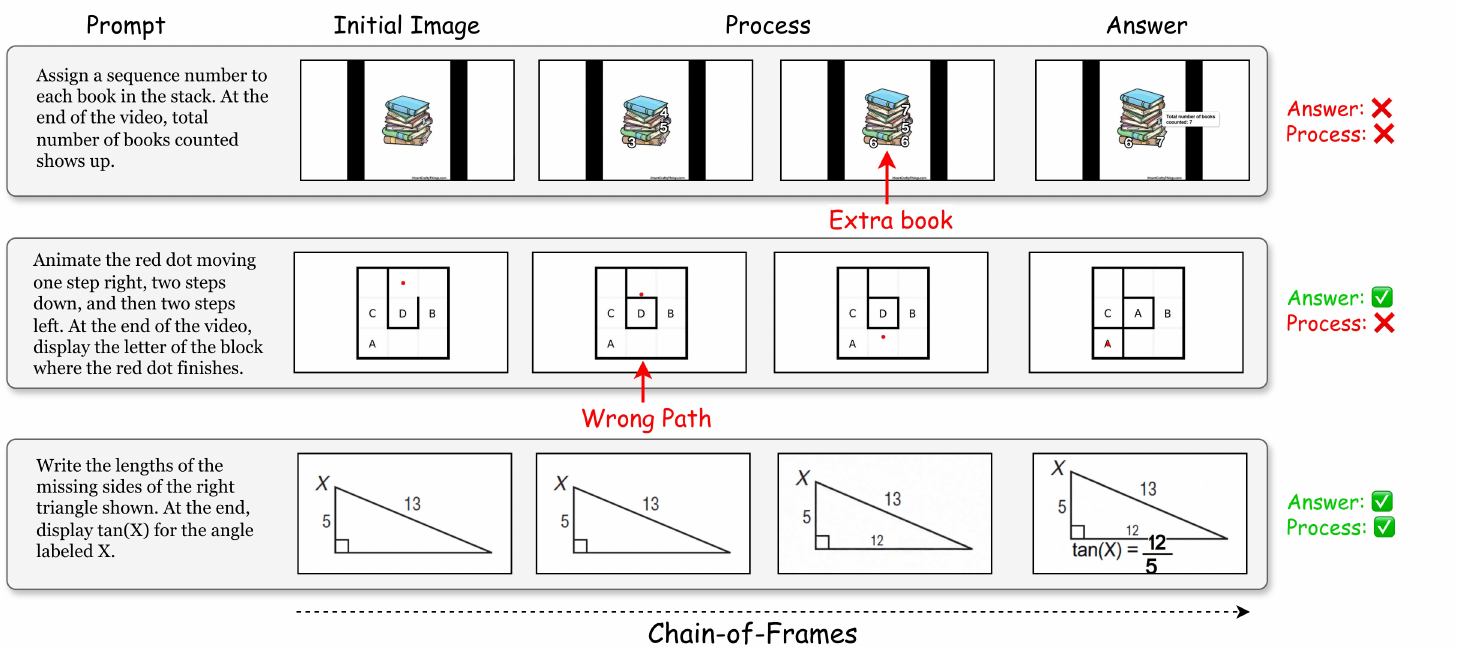}
    \caption{Illustration of reasoning via Chain-of-Frames. As shown, while models are capable of reaching the correct answer, they often generate erroneous intermediate frames (\eg ``Extra book'' or ``Wrong Path'').}
    \label{fig:empirical}
\end{figure*}

To better understand this phenomenon, researchers have introduced various Generative Video Reasoning (GVR) tasks. These tasks typically require a model to generate a video that completes a specific reasoning goal based on an initial image. While current evaluations show that video models possess promising reasoning potential, we identify several critical flaws in existing GVR assessment frameworks. A primary issue is that existing evaluation data and strategies provide insufficient focus on the procedural nature of video. Since a video consists of continuous frames, it possesses an inherent temporal and procedural attribute. However, existing evaluation strategies typically sample the last or the best-matched frame for assessment. This approach evaluates the correctness of the outcome while disregarding the validity of the reasoning process itself. According to our empirical study in Section~\ref{sec:empirical}, current video models produce \textbf{outcome-hacking}, where a model generates videos that reach a correct conclusion through an erroneous process.  Existing evaluation metrics based on single frame often misclassify these cases as correct, leading to an overestimation of the models' true reasoning capabilities.

To address these issues, we propose a process-aware evaluation paradigm for GVR. First, we introduce \textbf{VIPER} (VIdeo Process Evaluation for Reasoning tasks), a comprehensive benchmark consisting of 16 tasks across 6 domains, namely temporal, structural, symbolic, spatial, physics, and planning reasoning, comprising 309 items. To ensure a reliable evaluation of generated content, we propose a new metric, \textbf{Process-outcome Consistency} (POC$@r$). POC$@r$ takes frames sampled from the entire video at a sampling rate $r$ as input. For any given video, the POC score is determined by both its process consistency (PC) and outcome consistency (OC). OC requires at least one frame to fulfill the explicit task goal, while PC requires all sampled frames to adhere to the task's process-level constraints. A video is considered correct if and only if it fulfills both OC and PC. In practice, we follow the VLM-as-Judge paradigm to conduct these evaluations, utilizing a hierarchical rubric designed to facilitate accurate and extensible assessment.

Our experimental results demonstrate that even state-of-the-art video models only achieve approximately 20\% POC@1.0, while exhibiting an outcome-hacking rate of around 30\%. These findings indicate that a significant gap remains before current video models can be considered truly generalizable visual reasoners, particularly regarding the consistency between their processes and outcomes. We also investigate the effects of test-time scaling, a prevailing technique in language reasoning, on video models and find it yields notable performance improvements. Furthermore, we investigate the impact of the sampling rate $r$ on evaluation outcomes. Finally, we verify the reliability of our evaluation method through human-model alignment checks and summarize several common reasoning failure modes in existing models.

%% file: latex/2_empirical.tex
\begin{table}[t]
\centering
\begin{tabular}{lccc}
\toprule
\textbf{Model} & \textbf{Acc}$_{\text{last}}$ & \textbf{Acc}$_{\text{process}}$& \textbf{Hack} \\
\midrule
Veo 3.1  & 66.0 & 30.0 & 36.0 \\
Sora 2 & 70.0 & 24.0 & 46.0 \\
\bottomrule
\end{tabular}
\caption{Outcome-hacking analysis on video models.}
\vspace{-5px}
\label{tab:outcome-hacking}
\end{table}

\section{Outcome-hacking in Video Reasoning} \label{sec:empirical}
Existing GVR evaluation methods often rely on a single-frame strategy, typically assessing video quality based on either the last frame or the frame that best matches the final answer~\cite{think-with-video}. However, this approach may overlook process-level errors that occur in intermediate frames of the generated video. As a result, videos that exhibit flawed reasoning but reach the correct final result could be incorrectly judged as successful. We refer to this issue as \textbf{Outcome-hacking}. In this section, we conduct an empirical study on the outcome-hacking phenomenon in GVR tasks.

\subsection{Experiment Design}
To systematically examine this issue, we design an empirical study using a diagnostic dataset consisting of 50 task examples from existing GVR benchmarks~\cite{think-with-video, chen2025tivibench, zhou2025mira}. Specifically, we intentionally choose examples where the process is crucial to the final answer (\eg math problems or trajectory tracking). We then evaluate two advanced video models, Veo 3.1~\cite{veo3} and Sora 2~\cite{sora2}, by sampling one video per task. The evaluation is conducted using two metrics: (1) \textbf{Acc$_{\text{last}}$}: Evaluate the correctness of the video based on the final frame; (2) \textbf{Acc$_{\text{process}}$}: Evaluate the correctness of the video based on the entire video. We also report the discrepancy between these two metrics as the Hacking Rate, which highlights the degree of outcome-hacking present in the models.

\begin{figure*}[t]
    \centering
    \includegraphics[width=\linewidth]{images/overview.pdf}
    \caption{Overview of VIPER. VIPER consists of 16 tasks from 6 domains focusing on different reasoning abilities.}
    \vspace{-5px}
    \label{fig:overview}
\end{figure*}

\subsection{Empirical Results}
As illustrated in Table~\ref{tab:outcome-hacking}, both models achieve decent Acc$_{\text{last}}$ scores (66\% for Veo 3.1 and 70\% for Sora 2). Based solely on these results, it seems that video models have good reasoning capabilities. However, when the evaluation scope is expanded from the single last frame to the entire video, the accuracy drops significantly, with both models scoring below 30\%. Consequently, the hacking rate of Veo 3.1 reaches 36\%, while Sora 2's hacking rate is even higher, at 46\%. This result reflects that current video models are still far from being reliable visual reasoners, and it also motivates us to propose a more process-aware evaluation benchmark and metrics for more robust GVR evaluation.

%% file: latex/3_benchmark.tex
\section{The Proposed Benchmark: VIPER}
Inspired by the insights from our empirical study on outcome-hacking in GVR tasks, we propose a more process-aware evaluation benchmark, VIPER. We begin by formulating the generative video reasoning tasks, followed by an overview of the evaluation dimensions and the data collection pipeline.

\subsection{Generative Video Reasoning}
Generative Video Reasoning (GVR) involves generating a video that starts from a static visual context and fulfills a dynamic task. Given a video generation model $\mathcal{M}$, an initial image input $I$, and a task prompt $p$, a GVR task can be formulated as:
\begin{equation}
    V = \{f_1, f_2, \dots, f_n\} = \mathcal{M}(I, p),
\end{equation}
where $V$ represents the generated video consisting of frames $f_i$.

Notably, we define $p$ as a process-dependent task, where the validity of reasoning is determined by the joint consistency of the entire sequence $\{f_1, \dots, f_n\}$, rather than the correctness of any individual frame. Unlike tasks solvable by static image processing (\eg segmentation or super-resolution), a valid GVR task requires meaningful temporal evolution. Specifically, $p$ should describe a task that is either: 

\begin{itemize}
    \item \emph{Decomposable}, where the task involves a multi-step logical process, with intermediate frames serving as necessary milestones toward the solution (\eg path planning in a maze).
    \item \emph{Dynamic}, where the goal is to model the correct evolution of the scene, with the process itself representing the result (\eg predicting object collisions).
\end{itemize}

Under this definition, the task prompt $p$ can be decomposed into two components: 
\noindent
\begin{enumerate}
    \item \emph{Explicit target} ($t$): The outcome-oriented target explicitly stated in the prompt (\eg ``solve the maze'').
    \item \emph{Implicit constraints} ($C$): Rules not directly specified in the prompt but inherent to the task (\eg ``do not cross the walls of the maze'').
\end{enumerate}
A valid video $V$ must satisfy $t$ while strictly adhering to $C$ throughout all frames $f_i$.

\subsection{Evaluation Dimension}
To comprehensively evaluate the reasoning capabilities of video generation models, we propose six evaluation dimensions that focusing different aspect. Each dimension contains multiple tasks, as illustrated in Figure~\ref{fig:overview}. The detailed introduction of each task is presented in the Appendix~\ref{app:task_intro}.

\paragraph{Temporal Reasoning.} Tasks from this domain focus on evaluating the model's ability to accurately track and represent the dynamic evolution of objects over time. We design the following tasks: (1) \emph{Object Moving}: simulate the movement of simple geometric shapes to the target position. (2) \emph{World Memory}: simulate zooming in and out of an image while keeping the content unchanged.

\paragraph{Structural Reasoning.} Tasks from this domain focus on highly structured scenarios that require the model to follow strict, predefined rules and constraints. We include the following reasoning tasks: (1) \emph{Maze}: navigate from the start to the goal without violating maze constraints. (2) \emph{Checkmate-in-One}: find a single legal move that delivers checkmate under standard chess rules. (3) \emph{Tic-Tac-Toe}: choose a legal move to win or block the opponent under the game rules. (4) \emph{Sudoku}: fill the grid so that all rows, columns, and subgrids have no duplicated digits.

\paragraph{Symbolic Reasoning.} Tasks from this domain emphasize the model's ability to reason using symbolic information (\eg text tokens) within the generated video. We collect a variety of tasks: (1) \emph{Math Reasoning}: illustrate the process of solving a math problem. (2) \emph{Knowledge Reasoning}: solve problems based on world knowledge. (3) \emph{Multimodal Reasoning}: solve problems involving both text and image information.

\paragraph{Spatial Reasoning.} Tasks from this domain evaluate the model's ability to understand and reason about spatial relationships, geometric transformations, and the relative positioning of objects. We collect the following tasks: (1) \emph{Dice Rolling}: simulate the rolling of a die along a specified trajectory. (2) \emph{Cube Rotation}: simulate the rotation of a composite structure composed of multiple cubes. (3) \emph{Image Puzzle}: reassemble a set of shuffled sub-images into their original configuration.

\paragraph{Physics Reasoning.} This domain evaluates the model's ability to generate videos that adhere to real-world physical principles. We design two types of tasks: (1) \emph{Physics Experiment}: simulate experimental scenarios that demonstrate specific physical laws. (2) \emph{Physics Game}: predict the motion and state changes of objects in the game scenes involving object collisions and interactions.

\paragraph{Planning Reasoning.} This domain focuses on tasks that require the model to generate videos demonstrating the completion of multi-step complex tasks. It evaluates the model's ability to decompose complex tasks into individual steps and correctly execute each one. We design (1) \emph{Object Manipulation}: manipulate object within an embodied environment (2) \emph{Navigation}:  generate a correct, multi-step path based on a map.

\subsection{Data Collection}
Each sample in our dataset is formulated as a quadruple $\{i, p, C, g\}$, where $i$ represents the input image, $p$ denotes the task prompt, $C$ encapsulates the process constraints, and $g$ serves as the ground-truth reference (text, images, or videos). In total, we collect 309 items, detailed statistics are presented in Appendix~\ref{app:statistics}.

\paragraph{Input Image $i$.} Input images are curated from diverse sources, including web collection, Python programs, advanced image generation models, and selection from existing datasets.

\paragraph{Task Prompt $t$.} Each task prompt comprises three components: (1) \emph{Image caption}: A brief description of the visual content, which provides context and outlines the task rules for the model; (2) \emph{Task definition}: A description of the task objective (\eg ``a red line slowly traversing from the maze start point to the goal''); and (3) \emph{Style constraints}: Restrictions on the video generation style (\eg ``static camera'' or ``no zoom'').

\paragraph{Process Constraints $C$.} Although not explicitly stated in the task prompt, these constraints are inherent requirements for a valid solution video. For instance, in symbolic reasoning tasks, the model must demonstrate a correct problem-solving process rather than merely producing a correct final answer. We include these information to assist process-level evaluation.

\paragraph{Reference $g$.} To facilitate precise evaluation, we provide a reference $g$ representing the ground-truth process or outcome. The format of $g$ varies by task: it is sampled from the ground-truth video if available; otherwise, it is an image of the solved state. For symbolic reasoning tasks, $g$ comprises the textual reasoning steps and the final answer.

%% file: latex/4_evaluation.tex
\section{Process-centric Video Evaluation}
To enable accurate evaluation of the GVR task, we propose a novel metric, POC$@r$, which uniformly sampling multiple frames across the video and evaluate videos from both outcome and process level. Furthermore, to ensure scalable and reproducible evaluation, we adopt the VLM-as-Judge pradigm, combined with a hierarchical rubric tailored to various domain-specific tasks.

\subsection{Process-outcome Consistency}
As previously discussed, evaluating GVR tasks based solely on a single frame can lead to severe outcome-hacking, resulting in an inflated estimation of the model's capabilities. To address this, we propose \textbf{Process-outcome Consistency (POC)}, a new metric designed to comprehensively assess the correctness of the generated video at both outcome- and process-level.

Formally, let $V = \mathcal{M}(i, t)$ denote the video generated by model $\mathcal{M}$. Instead of selecting the last or best frame for evaluation, POC uniformly samples multiple frames from $V$ at a frame rate of $r$ (denoted as $\text{POC}@r$), obtaining the frame set $\hat{V}_r = \{ f_k \}_{k=1}^{N}$, where $N$ is the total number of sampled frames. To determine whether a frame adheres to the text condition, we employ a VLM as the judge model $\mathcal{J}$. Specifically, POC$@r$ is formulated as:
\begin{equation}
\begin{aligned}
\text{POC}@r &= \mathcal{J}(\hat{V}_r, t, C, g) 
= \text{OC}@r \land \text{PC}@r,\\
\text{OC}@r 
&= \mathbbm{1}\!\left[\exists f \in \hat{V}_r, \; f \sim t\right], \\
\text{PC}@r
&= \mathbbm{1}\!\left[\forall f \in \hat{V}_r,\; f\sim C\right]. \\
\end{aligned}
\end{equation}
As shown in the formulation, the judge $\mathcal{J}$ evaluates outcome consistency (OC$@r$) and process consistency (PC$@r$) independently, taking their logical conjunction as the final POC score. Specifically, OC$@r$ requires that there exists at least one frame in $\hat{V}_r$ that satisfies the task prompt $t$, whereas PC$@r$ requires that every frame within $\hat{V}_r$ must adhere to the process constraints $C$. This formulation ensures that a generated video is deemed correct under POC if and only if it fulfills the task requirements via a valid process, thereby effectively mitigating the issue of outcome-hacking.

\begin{table*}[ht]
    \centering
    \scalebox{0.9}{
    \begin{tabular}{lccccccccccc}
    \toprule
    \multirow{2.5}{*}{\textbf{Model}} &\multirow{2.5}{*}{\textbf{Temporal}} & \multirow{2.5}{*}{\textbf{Structural}} & \multirow{2.5}{*}{\textbf{Symbolic}} & \multirow{2.5}{*}{\textbf{Spatial}} & \multirow{2.5}{*}{\textbf{Physics}} & \multirow{2.5}{*}{\textbf{Planning}} & \multicolumn{3}{c}{\textbf{Overall}} \\
    \cmidrule{8-10}
    &&&&&&& OC$\uparrow$ & Hack $\downarrow$ & POC $\uparrow$\\
    \midrule
    \rowcolor{gray!20}
    \multicolumn{10}{c}{\emph{Open-source Models}} \\
    \midrule
    Wan 2.2 &4.0	&11.3	&0.0&	1.3	&8.4&	21.2&	26.6&	27.9&7.7\\
    Hunyuan 1.5  & 8.4&	5.0	&0.0&	2.7	&11.1&	21.1	&27.5	&	\textbf{19.4}&8.1 \\
    \midrule
    \rowcolor{gray!20}
    \multicolumn{10}{c}{\emph{Proprietary Models}} \\
    \midrule
    Seedance 1.5 &5.6	&5.6&	0.0&	5.3	&6.5&	33.8&	41.1&31.6 &	9.5	\\
    Wan 2.6 & \textbf{26.2} &	23.8&	0.0&	\textbf{12.0}&	13.9	&35.6&	42.6&	24.0&	18.6\\
    Veo 3.1&  22.2	&20.0&	13.3&	10.7	&\textbf{14.7}&	\textbf{41.0}&\textbf{56.1} & 35.8 & 20.3\\
    Sora 2& 10.4&	\textbf{42.5}&	\textbf{58.3}&	4.0&9.4&	15.1&47.0& 23.7 &\textbf{23.3}\\
    \bottomrule
    \end{tabular}}
    \caption{Performance of mainstream video models arcoss 16 tasks within VIPER. We report the POC$@$1.0 for each domain and summarize the average OC, POC and hacking rate. Best performance is \textbf{bolded}.}
    \label{tab:main}
\end{table*}

\subsection{Hierarchical Rubric}
To facilitate accurate consistency judgment, we design a hierarchical rubric for $\mathcal{J}$. Specifically, the rubric is structured into three distinct levels. Specific examples are provided in Appendix~\ref{app:rubric}.

\paragraph{System Prompt.} At the top tier, system prompt defines the task context, evaluation criteria, and methodology. Adopting the popular test-time scaling paradigm~\cite{deepseekr1}, the system prompt requires the model to first generate the analysis of process and outcome consistency within \texttt{<think>} tags, and subsequently provide the decision for PC, OC and POC within \texttt{<answer>} tags.

\paragraph{Domain Introduction.} In the middle, domain introduction provides a concise domain overview, highlighting the specific evaluation focus within the field. Crucially, all tasks belonging to the same domain share this unified introduction.

\paragraph{Task Constraints.} At the instance level, task constraints imposes fine-grained evaluation criteria, which are reorganized from the  process constraints $C$ and the reference $g$ collected previously.

Collectively, this hierarchical rubric covers from macro-level task definitions to micro-level requirements, providing comprehensive evaluation guidance. Moreover, this design is highly scalable. To support new tasks or domains, we only need to add the corresponding constraints and introductions.

%% file: latex/5_experiment.tex
\section{Experiment}

\subsection{Experiment Setup}

\paragraph{Baseline Models.} We evaluate mainstream proprietary
and open-source video generation models, including Sora 2~\cite{sora2}, Veo 3.1~\cite{veo3}, Seedance 1.5 pro~\cite{chen2025seedance}, Wan 2.2/2.6~\cite{wan2025wan} and HunyuanVideo-1.5~\cite{hunyuan}.

\paragraph{Implementation.}
By default, we generate one video for each prompt and take POC$@$1.0 as the main metric. And we select the GPT-5~\cite{gpt5} as the judge model $\mathcal{J}$ due to its advanced multimodal understanding capabilities.

\subsection{Experiment Result} We present the main experiment results in Table~\ref{tab:main}, from which we summarize following observations. Firstly, all models struggle with VIPER, with POC$@$1.0 scores consistently below 30\%. Sora 2 performs the best, achieving a score of 23.3\%, followed by Veo 3.1 and Wan 2.6, scoring 20.3\% and 18.6\%. The remaining models fail to surpass a score of 10\%. This indicates that current video generation models are not yet capable of functioning as general visual reasoners.

Furthermore, severe outcome-hacking is observed across all models. Notably, Veo 3.1 experiences the most severe hacking, with a rate of 35.8\%. This suggests that, while these models possess the necessary knowledge for fulfill reasoning goals, they struggle to faithfully represent the reasoning process through video frames. As such, improving process-level reasoning is a crucial step toward advancing model performance.

Lastly, the models demonstrate varying strengths across different tasks. Symbolic reasoning, in particular, proves to be a major challenge for most models, which struggle to generate accurate and legible text. However, Sora 2 excels in this domain, achieving a POC of 58.3\%, which highlights its robust text rendering and reasoning capabilities. In contrast, for tasks like planning and physics which are more closely tied to real-world scenarios, Veo 3.1 outperforms Sora 2, underscoring its superior capabilities in real-world physics simulation.

\begin{tcolorbox}[
    colback=gray!5!white, colframe=gray!55!black, 
    arc=10pt,           
    boxrule=1.5pt,      
    boxsep=5pt,         
    left=2pt, right=2pt, top=2pt, bottom=2pt 
]
    \textbf{Takeaway 1 } Current video models are far from being universal visual reasoners, often reaching correct goal via invalid trajectories.
\end{tcolorbox}

\begin{table*}[ht]
    \centering
    \scalebox{0.95}{
        \begin{tabular}{llccccccc}
            \toprule
            \textbf{Model} & \textbf{Metric} & \textbf{Temporal} & \textbf{Structural} & \textbf{Symbolic} & \textbf{Spatial} & \textbf{Physics} & \textbf{Planning} & \textbf{Overall} \\
            \midrule
            
            \multirow{3}{*}{Veo 3.1} 
            & Pass$@$1 & 22.2 & 20.0 & 13.3 & 10.7 & 14.7 & 41.0 & 20.3 \\
            & Pass$@$4 & 33.4 & 35.0  & 21.7 & 23.0   & 25.8 & 67.0   & 34.3 \\
            & Pass$@$8 & 37.4 & 55.0 & 31.7 & 25.7 & 35.7 & 73.4 & 43.2 \\
            \midrule
            
            \multirow{3}{*}{Sora 2} 
            & Pass$@$1 & 10.4 & 42.5 & 58.3 & 4.0  & 9.4  & 15.1 & 23.3 \\
            & Pass$@$4 & 34.2 & 73.8 & 91.7 & 14.7 & 38.5 & 23.8 & 46.1 \\
            & Pass$@$8 & 58.9 & 75.0 & 95.0 & 26.1 & 49.3 & 47.0 & 58.6 \\
            
            \bottomrule
        \end{tabular}
    }
    \caption{Effect of test-time scaling. We calculate Pass$@$1, Pass$@$ and Pass$@$8 for Veo 3.1 and Sora2 with POC$@$1.0.}
    \vspace{-5px}
    \label{tab:test-time-scaling}
\end{table*}

\subsection{Further Analysis}
\paragraph{Test-time Scaling.} We investigate whether test-time scaling, effective in language reasoning, similarly benefits video reasoning. Pass$@$$k$ is a widely used metric for evaluating language models, with recent studies indicating that performance can be significantly enhanced by increasing the number of samples $k$~\cite{passk, wang2023self-consistency}. To validate this in the GVR tasks, we sample multiple videos on VIPER using Sora 2 and Veo 3.1 and reporting Pass$@$1, 4, and 8 with POC$@$1.0 in Table~\ref{tab:test-time-scaling}. As illustrated, scaling the sampling times notably improves performance. For instance, the overall accuracy of Veo 3.1 increases from 20.3\% to 43.2\%, particularly in challenging domains such as symbolic reasoning. However, performance gains saturate as $k$ increases. Domains such as spatial and physics reasoning remain challenging, indicating that while test-time scaling provides a notable boost, it cannot fully compensate for reasoning limitations in current video models.

\begin{table}[t]
    \centering
    \begin{tabular}{lrccc}
    \toprule
     \textbf{Model} &  $r$   & \textbf{OC}$\uparrow$ & \textbf{PC}$\uparrow$&  \textbf{POC} $\uparrow$  \\
     \midrule
    \multirow{3}{*}{Veo 3.1}         & 0.5 & 61.5 & 33.7 & 28.2 \\
         & 1.0 & 59.7 & 21.8 & 20.3 \\
         & 2.0 & 58.9 & 16.5 & 15.4 \\
         \midrule
    \multirow{3}{*}{Sora 2}        & 0.5 & 47.1 & 35.4 & 28.0 \\
         & 1.0 & 51.8 & 31.0 & 23.3 \\
         & 2.0 & 48.5 & 32.0 & 21.9 \\
     \bottomrule
    \end{tabular}
    \caption{Ablation on the sampling rate $r$.}
    \label{tab:sampling-rate}
\end{table}

\begin{tcolorbox}[
    colback=gray!5!white, colframe=gray!55!black, 
    arc=10pt,           
    boxrule=1.5pt,      
    boxsep=5pt,         
    left=2pt, right=2pt, top=2pt, bottom=2pt 
]
    \textbf{Takeaway 2 } While test-time scaling improves video model performance, it fails to resolve fundamental reasoning bottlenecks.
\end{tcolorbox}

\paragraph{Ablation on Sampling Rate.} We investigate the impact of the sampling rate $r$ on the POC$@r$ metric. Intuitively, a higher $r$ entails denser frame sampling, thereby imposing a more rigorous criterion for POC assessment. As illustrated in Table~\ref{tab:sampling-rate}, the model's performance exhibits a consistent decline as $r$ increases. While a higher sampling rate enhances evaluation rigor, it simultaneously extends the input sequence for the judge model, leading to increased computational overhead and slower evaluation. To strike an optimal balance between assessment rigor and efficiency, we adopt $r=1.0$ as the default setting. Nevertheless, users may increase $r$ for more rigorous evaluation.

\begin{figure}[t]
    \centering
    \includegraphics[width=\linewidth]{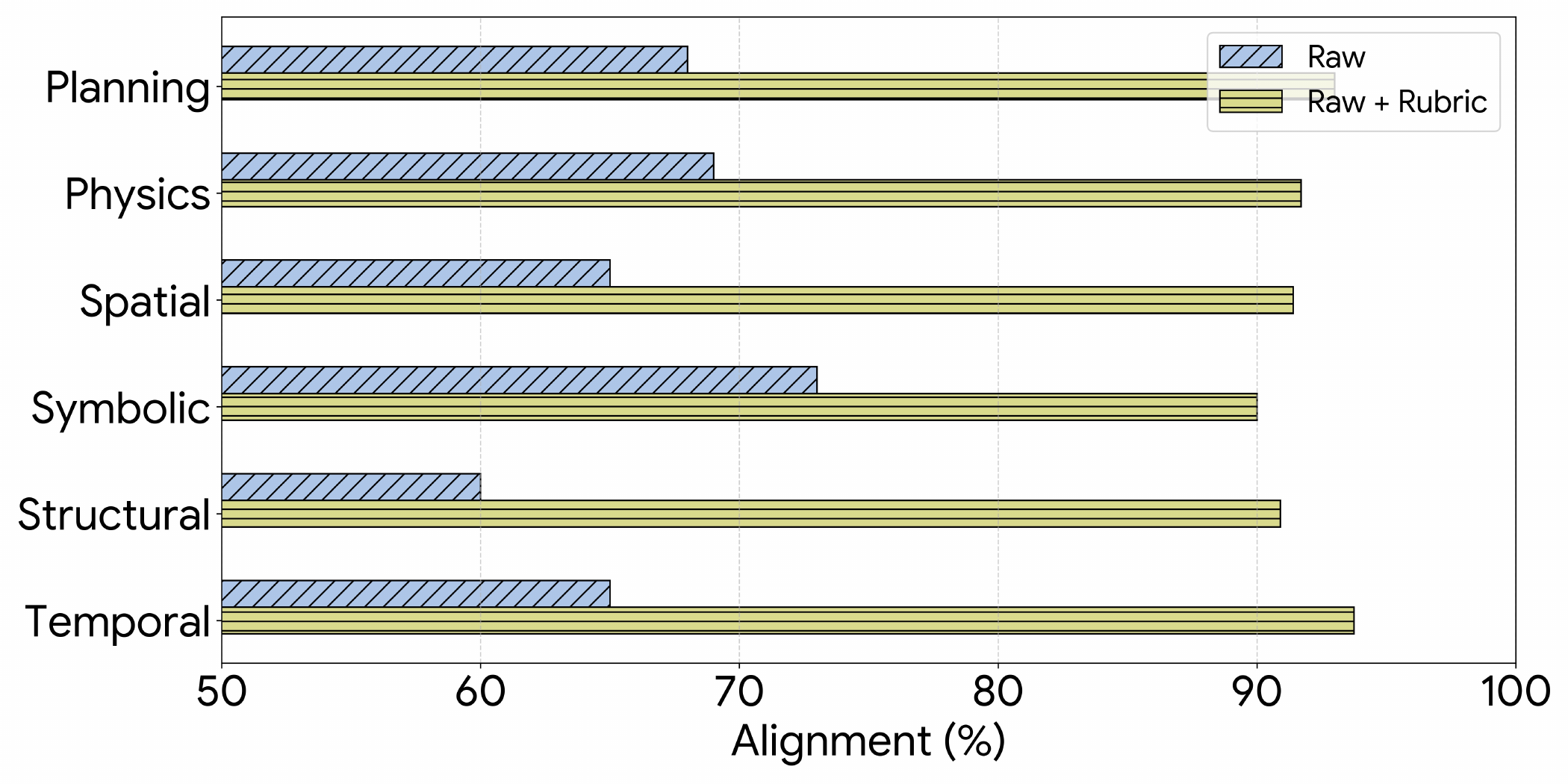}
    \caption{Human-model alignment under combinations of raw input and hierarchical rubric.}
    \vspace{-5px}
    \label{fig:alignment}
\end{figure}

\paragraph{Human-model Alignment.}
To further validate the robustness of our evaluation paradigm, we compare the evaluation results from the judge model with those from human experts. As illustrated in Figure~\ref{fig:alignment}, by providing the judge model with our hierarchical rubric, its alignment with human evaluators generally exceeds 90\%, whereas the raw evaluation achieves only about 60\%. These results demonstrate the effectiveness of our rubric.

\begin{tcolorbox}[
    colback=gray!5!white, colframe=gray!55!black, 
    arc=10pt,           
    boxrule=1.5pt,      
    boxsep=5pt,         
    left=2pt, right=2pt, top=2pt, bottom=2pt 
]
    \textbf{Takeaway 3 } A hierarchical rubric is essential for reliable VLM-based evaluation, bridging the gap between automated metrics and human judgment where unstructured prompts fail.
\end{tcolorbox}

\begin{figure*}[t]
  \centering
  \includegraphics[width=\linewidth]{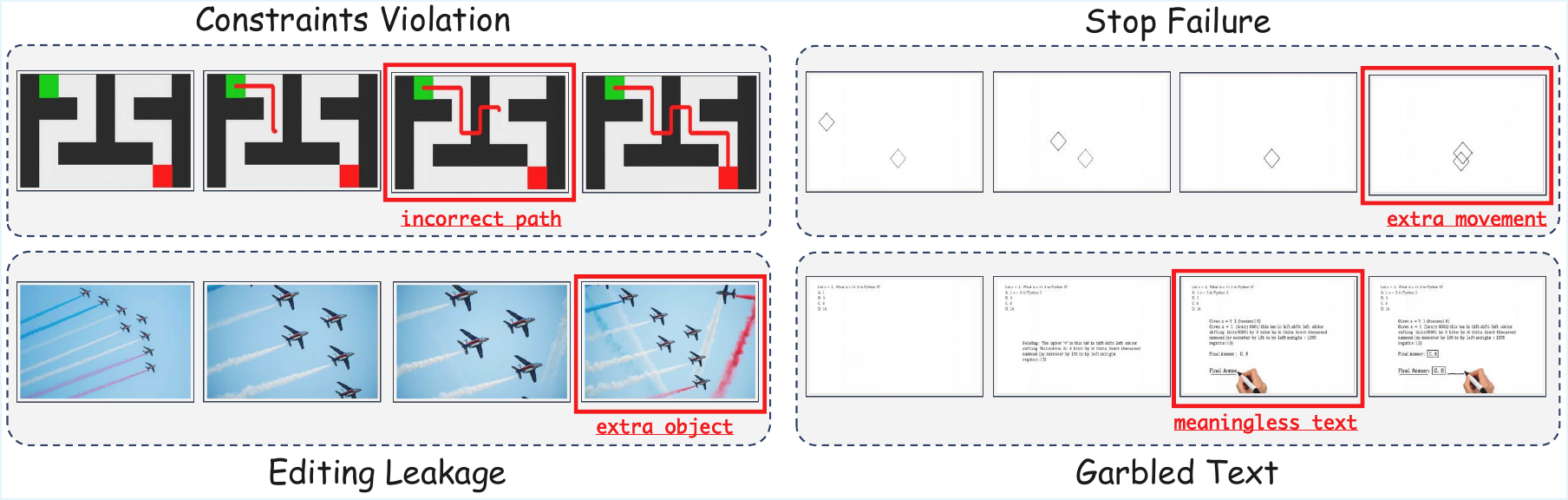}
  \caption{Representative failure patterns observed from VIPER.}
  \label{fig:fail_case}
\end{figure*}

\begin{table*}[h]
\centering
\begin{tabular}{lccccccc}
\toprule
\textbf{Model} & \textbf{Temporal} &\textbf{ Structural} & \textbf{Symbolic} &\textbf{ Spatial} & \textbf{Physics} & \textbf{Planning} & \textbf{Overall}\\
\midrule
Veo 3.1 & 22.2 & 20.0 & 13.3 & 10.7 & 14.7 & 41.0 & 20.3 \\
+ $C$ & 22.2 & 31.3 & 13.3 & 14.7 & 17.2 & 37.6 & 22.7 \\
\bottomrule
\end{tabular}
\caption{Performance Comparison of Veo 3.1 without and with constraints $C$ as input.}
\vspace{-7px}
\label{tab:constraints}
\end{table*}

\subsection{Failure Patterns}
Upon observing generation results from VIPER, we identify several representative failure modes prevalent in current video models. We categorize these patterns below and illustrate specific cases in Figure~\ref{fig:fail_case}, aiming to provide insights for the future optimization of video generation models.

\paragraph{Constraints Violation.} We observe that models often struggle to identify implicit constraints within tasks, such as not violating the maze structure when solving the maze or adhering to the movement rules in a checkmate-in-one scenario. To address this, we investigate whether explicitly prompting these constraints can mitigate the issue. We evaluate Veo 3.1 by supplementing the input with explicit process constraints $C$. As shown in Table~\ref{tab:constraints}, providing this auxiliary information significantly enhances model performance. This result indicates that the failure to respect constraints is not attributed to a deficiency in the model's generation capabilities, but rather stems from an insufficient comprehension of the instructions.

\begin{tcolorbox}[
    colback=gray!5!white, colframe=gray!55!black, 
    arc=10pt,           
    boxrule=1.5pt,      
    boxsep=5pt,         
    left=2pt, right=2pt, top=2pt, bottom=2pt 
]
    \textbf{Takeaway 4 } The failures of video models are sometimes not due to a lack of fundamental generation abilities, but rather the difficulty in uncovering implicit intentions.
\end{tcolorbox}

\paragraph{Stop Failure.} Current video generation models typically produce outputs of fixed duration. However, reasoning tasks like checkmate-in-one often require short, precise sequences (\eg 1-2 seconds). Ideally, the model should maintain a static frame upon task completion. Ideally, the model should maintain a still frame after completing the task. Instead, current models frequently continue to generate content after fulfill the task, introducing unnecessary motion or hallucinations that disrupt the integrity of the reasoning process.

\paragraph{Editing Leakage.}
Unlike open-ended video generation, GVR tasks are highly structured and require precise, controlled editing. Models must strictly adhere to instructions without altering unrelated regions. However, we observe that current models struggle to define clear editing boundaries, often exhibiting ``leakage'' where background elements are altered, extra objects are introduced, or original task conditions are violated.

\paragraph{Garbled Text.}
The readability of text within the generated video is a crucial factor in determining reasoning quality, particularly in symbolic reasoning tasks where the model is expected to render the solution process in textual form. Currently, most models struggle to generate fluent and readable text passages, often producing garbled or incoherent glyphs. As a result, even if the reasoning outcome is correct, the video fails to provide a coherent and interpretable reasoning process.

%% file: latex/6_related_work.tex
\section{Related Work}
\subsection{Video Generation Models}
Recent advancements in diffusion-based Transformers and architecture optimization have significantly accelerated progress in video generation \citep{ho2022video, yang2024cogvideox, zheng2024open, jin2024pyramidal, blattmann2023align}. Modern video generation systems can now generate high-fidelity videos conditioned on both initial images and textual instructions. On the proprietary side, leading state-of-the-art models include Sora 2 \cite{sora2}, Veo 3.1 \cite{veo3}, Seedance 1.5 Pro \cite{chen2025seedance}, and Wan 2.6~\cite{wan2025wan}, all of which exhibit improved temporal consistency and enhanced ability to simulate complex physical interactions over extended time horizons. Concurrently, the open-source community has made significant strides, with models like Wan 2.2 \cite{wan2025wan} and HunyuanVideo \cite{hunyuan} achieving competitive performance. These models leverage architectural innovations, such as 3D VAEs and flow-matching Transformers, to enhance spatiotemporal modeling and generation quality.

\subsection{Generative Video Reasoning}
As video generation models evolve, their potential for completing reasoning tasks has gradually been uncovered. Research has shown that video generation models can generalize to many common visual tasks that they have not encountered during training, using a sequence of frames to represent the solving process, which is termed Chain-of-Frames (CoF) reasoning~\cite{wiedemer2025veo3-zeroshot}. Subsequent work has focused on evaluating and optimizing video models for generative video reasoning tasks. Some studies have designed empirical studies to showcase the reasoning capabilities of video models~\cite{guo2025mmecof, think-with-video}. Additionally, many efforts have proposed benchmarks covering various evaluation dimensions by collecting existing video data or manually constructing examples~\cite{chen2025tivibench, deng2025video, zhou2025mira, liu2025gen-vire, luo2025v-reasonbench}. Furthermore, some studies have made initial explorations into improving models' reasoning abilities through prompt optimization and supervised fine-tuning using correct video examples, yielding some promising results~\cite{chen2025tivibench, miniveo3reasoner}.

%% file: latex/7_conclusion.tex
\section{Conclusion}
In this work, we present VIPER, a comprehensive process-aware evaluation benchmark for Generative Video Reasoning (GVR). By introducing Process-outcome Consistency (POC$@r$), we address critical limitations in existing evaluation frameworks, which often overlook the process-level integrity of generated videos. Our experiments reveal that while state-of-the-art video models demonstrate some reasoning capabilities, they suffer from outcome-hacking, with performance on the VIPER falling below expectations. We identify key failure modes such as constraints violation, stop failure, and editing leakage, which impede the generation of accurate, reasoning-oriented videos. Despite the gap between current video models and true generalized visual reasoning remains substantial, we show that scaling sampling times and  incorporating process-level constraints can offer some improvements. Moving forward, it is clear that enhancing process consistency and bridging the reasoning gap will be essential for advancing video models toward becoming reliable general-purpose visual reasoners.

%% file: latex/8_appendix.tex
\section{Benchmark Statistics} \label{app:statistics}
As illustrated in Table~\ref{tab:benchmark_stats}, VIPER comprises a total of 309 samples. We curate most tasks to contain 10-25 samples, ensuring a balanced distribution to assess model performance across diverse scenarios. The Structural category accounts for the largest portion, followed by the Symbolic and Physics domains.
\begin{table}[h]
    \centering
    \scalebox{0.9}{
    \begin{tabular}{llcc}
        \toprule
        \textbf{Domain} & \textbf{Task} & \textbf{Count} & \textbf{Total} \\
        \midrule
        \multirow{2}{*}{Physics} & Experiment & 18 & \multirow{2}{*}{32} \\
         & Game & 14 & \\
        \midrule
        \multirow{2}{*}{Planning} & Navigation & 25 & \multirow{2}{*}{44} \\
         & Object Manipulation & 19 & \\
        \midrule
        \multirow{3}{*}{Spatial} & Block Rotate & 25 & \multirow{3}{*}{60} \\
         & Dice Rolling& 20 & \\
         & Image Restore & 15 & \\
        \midrule
        \multirow{4}{*}{Structural} & Chess & 20 & \multirow{4}{*}{70} \\
         & Maze & 20 & \\
         & Sudoku & 20 & \\
         & Tic-Tac-Toe & 10 & \\
        \midrule
        \multirow{3}{*}{Symbolic} & Knowledge & 20 & \multirow{3}{*}{60} \\
         & Math & 20 & \\
         & Multimodal & 20 & \\
        \midrule
        \multirow{2}{*}{Temporal} & Object Move & 25 & \multirow{2}{*}{43} \\
         & World Memory & 18 & \\
        \midrule
        \textbf{Sum} & & & \textbf{309} \\
        \bottomrule
    \end{tabular}}
    \caption{Detailed statistics of VIPER.}
    \label{tab:benchmark_stats}
\end{table}

\section{Task Introduction} \label{app:task_intro}
\subsection{Temporal Reasoning}

\TaskEntry{Object Moving.}
{This task requires the model to smoothly move a geometric shape to a certain position. Images are generated programmatically using Python.}
{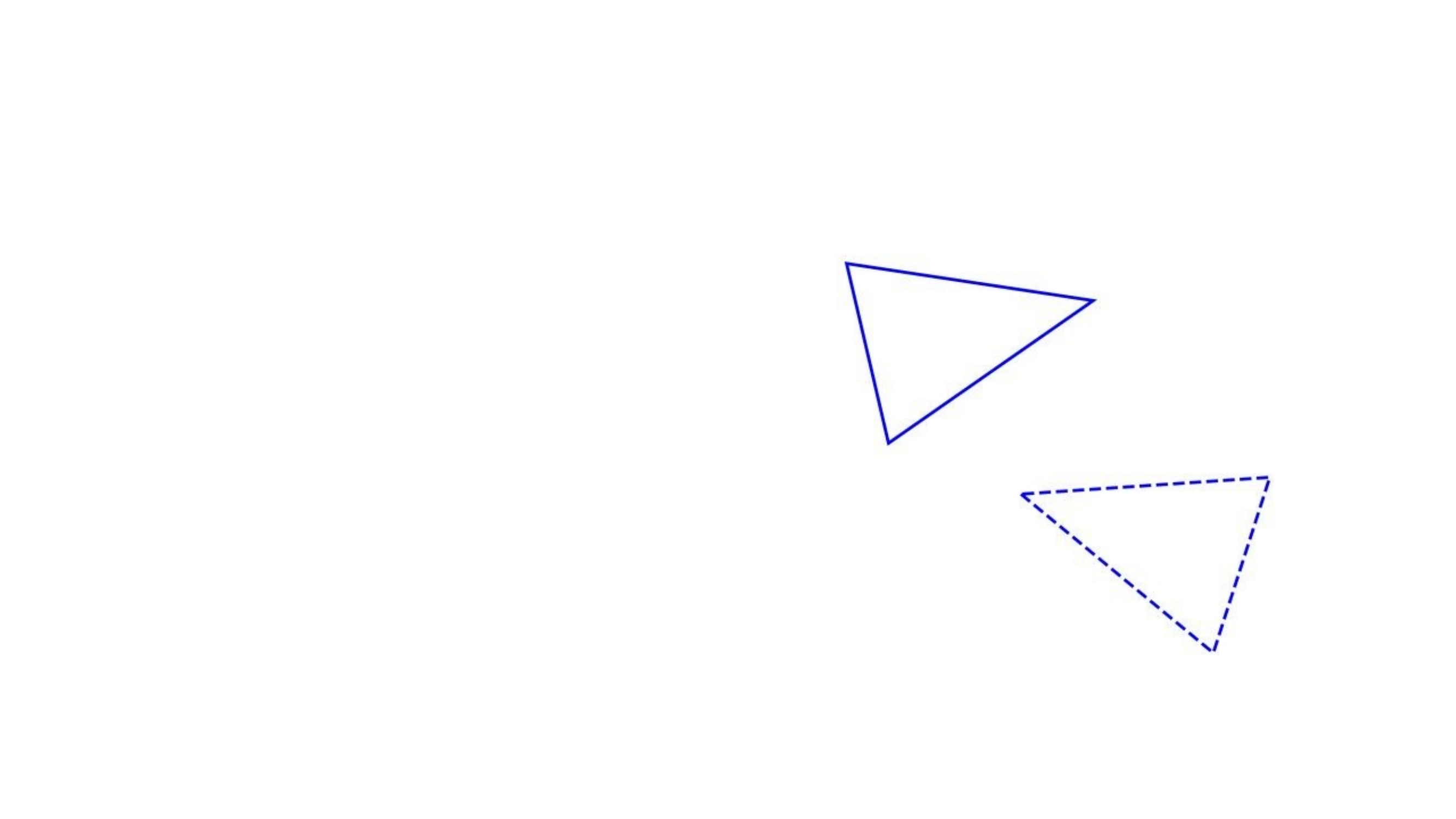}
{A clean white background featuring a blue triangle defined by a solid outline at top, and a target destination marked by a matching blue triangle with a dotted outline. The solid-outlined object travels in down, ending perfectly overlapping and aligning with the dotted target outline. The object moves by smoothly rotating clockwise while translating. High contrast, vector graphic style, locked camera, no zoom, consistent scale.}

\TaskEntry{World Memory.}
{This task requires the model to perform a centered zoom-in and then restore to the original view with visual elements unchanged. Images are collected from open-source web sources.\footnote{https://unsplash.com/}.}
{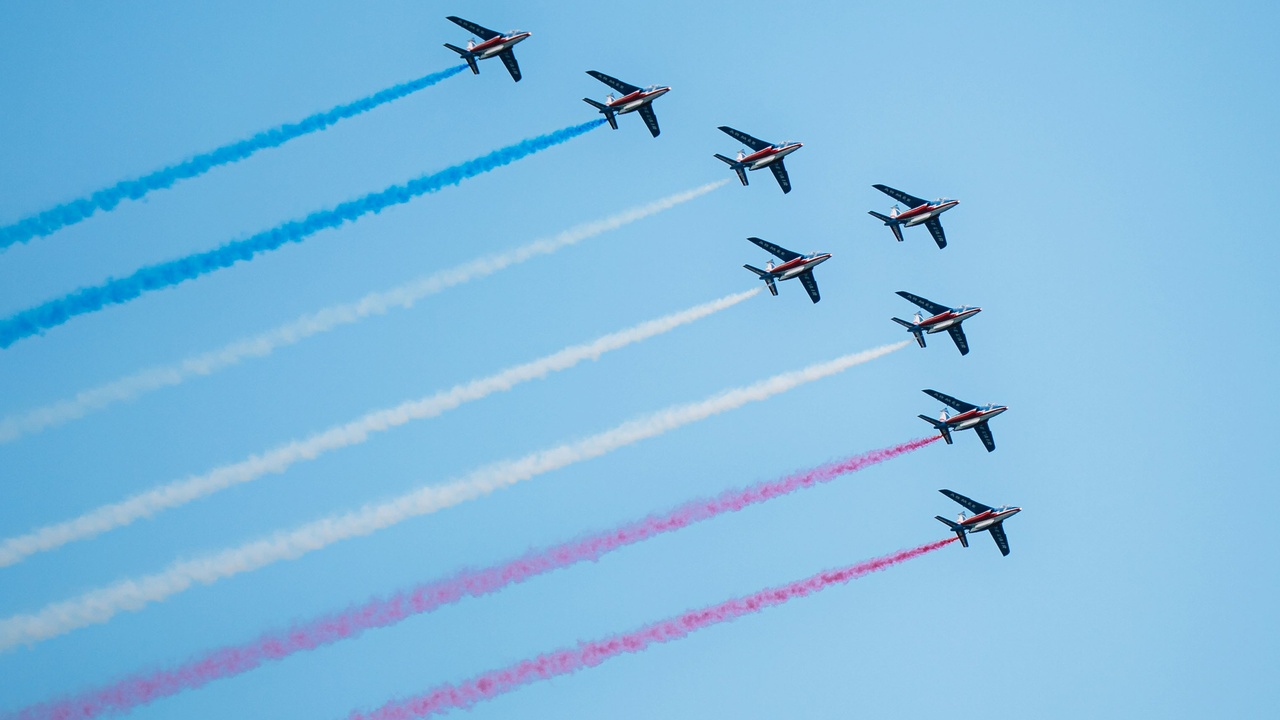}
{An image. Execute a slow, centered zoom-in to 2x magnification, hold the frame static, and then smoothly zoom out to the original field of view. The visual elements remain static and unchanged throughout the entire process. No pan, no dolly, no perspective shifts.}

\subsection{Structural Reasoning}

\TaskEntry{Sudoku.}
{This task requires the model to complete a partially filled Sudoku grid by correctly filling the empty cells with numbers. Images are generated programmatically using Python from valid Sudoku puzzles.}
{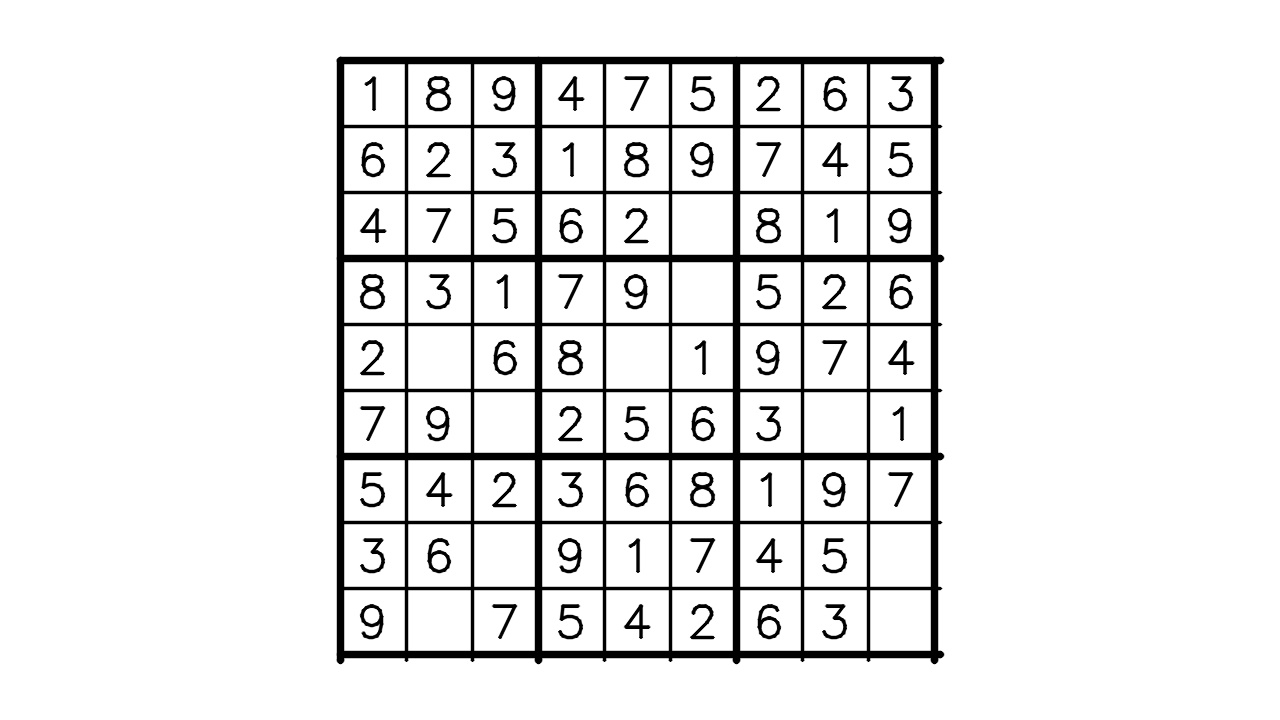}
{A clean, high-contrast image of a partially filled 9x9 Sudoku grid on a solid white background. Sovle the Sudoku, with the missing numbers (1-9) appearing in the correct empty cells. Top-down orthographic view, locked camera, no zoom, no pan, no perspective shifts, consistent scale.}

\TaskEntry{Tic Tac Toe.}
{This task requires the model to interpret a midgame Tic-Tac-Toe board and place in an appropriate empty cell according to the game state. Images are generated programmatically using Python from valid Tic-Tac-Toe configurations.}
{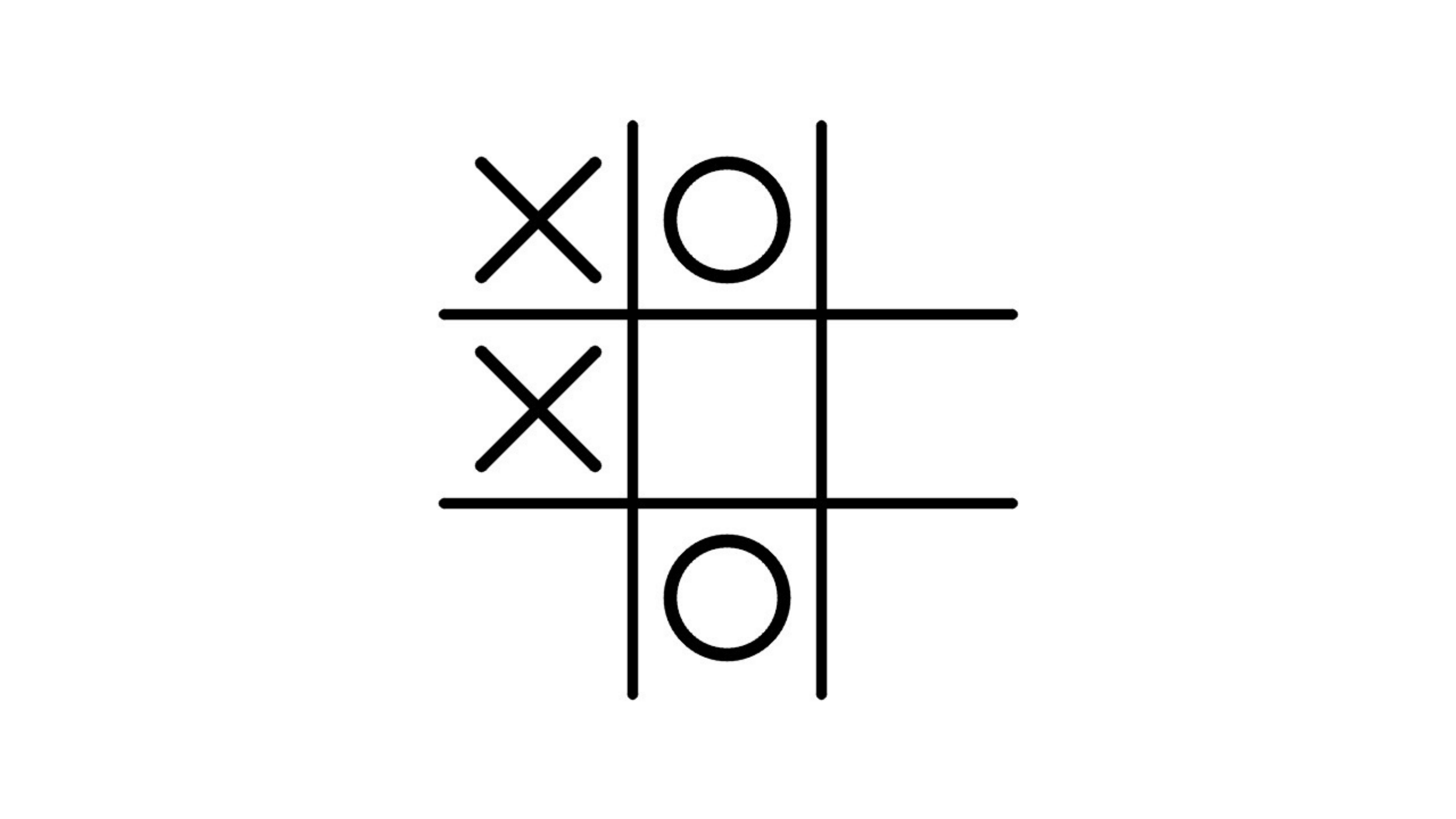}
{A static, top-down view of a minimalist 3x3 Tic-Tac-Toe grid on a clean background. The board is partially filled with existing 'X' and 'O' markers in a midgame state. A single 'X' is drawn into a strategic empty square, executing a logical move to block an opponent or complete a straight line of three. The grid lines and existing marks remain undisturbed. High contrast, vector graphic style, locked camera, no zoom, no dolly, no perspective shifts.}

\TaskEntry{Checkmate-in-One.}
{This task requires the model to interpret a chess position from an image and generate a single legal move that results in checkmate. Images are generated programmatically using Python from valid, solvable chess puzzles~\cite{bigbench}.}
{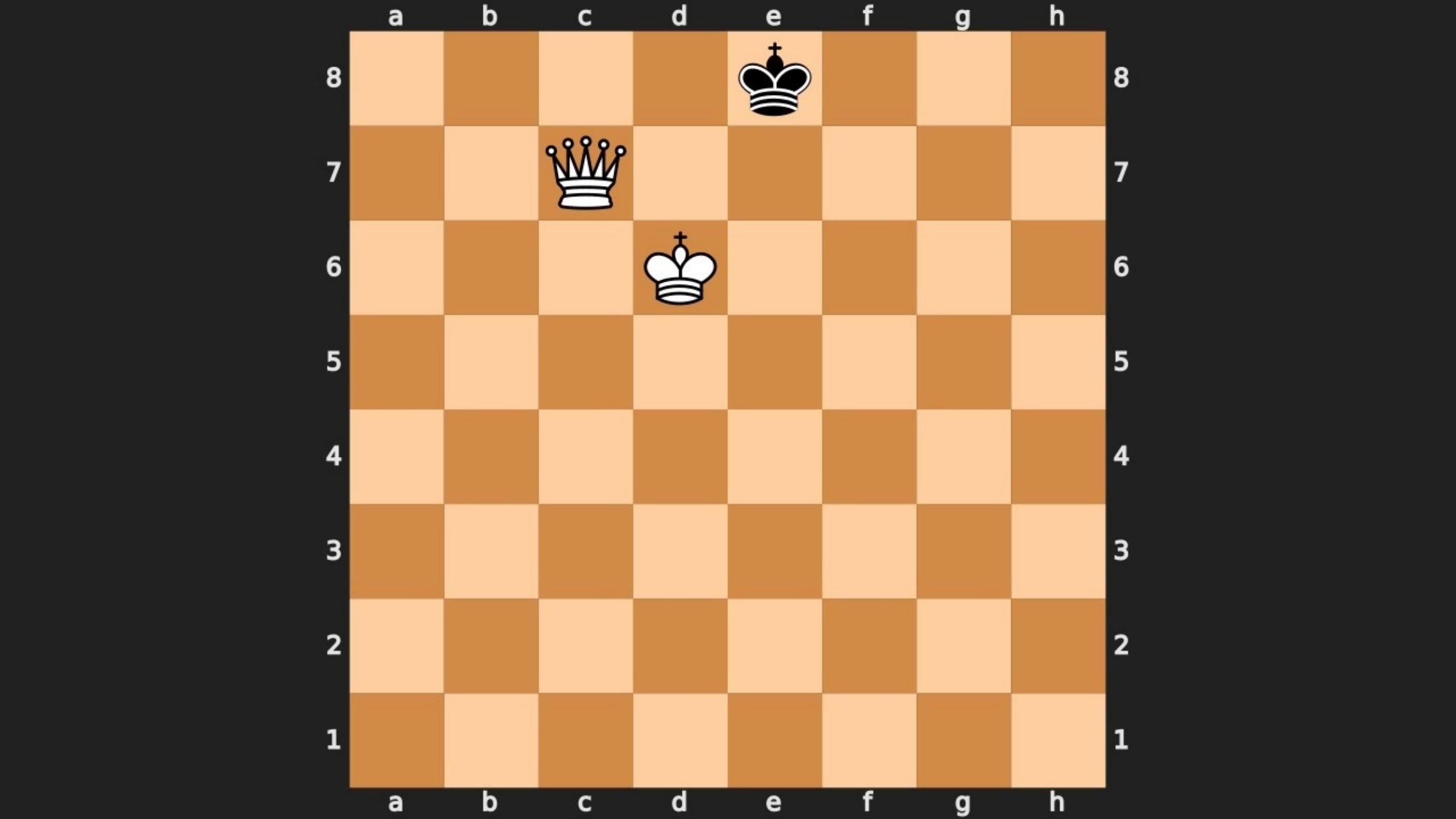}
{A static, top-down 2D view of a digital chess board. A single White piece glides smoothly for exactly one step to deliver checkmate to the Black King. Static camera, no pan, no zoom, no dolly, no perspective shifts.}

\TaskEntry{Maze.}
{This task requires the model to draw a continuous path through a 2D maze from a given start point to an end point without crossing any walls. Images are generated programmatically using Python from randomly generated solvable mazes.}
{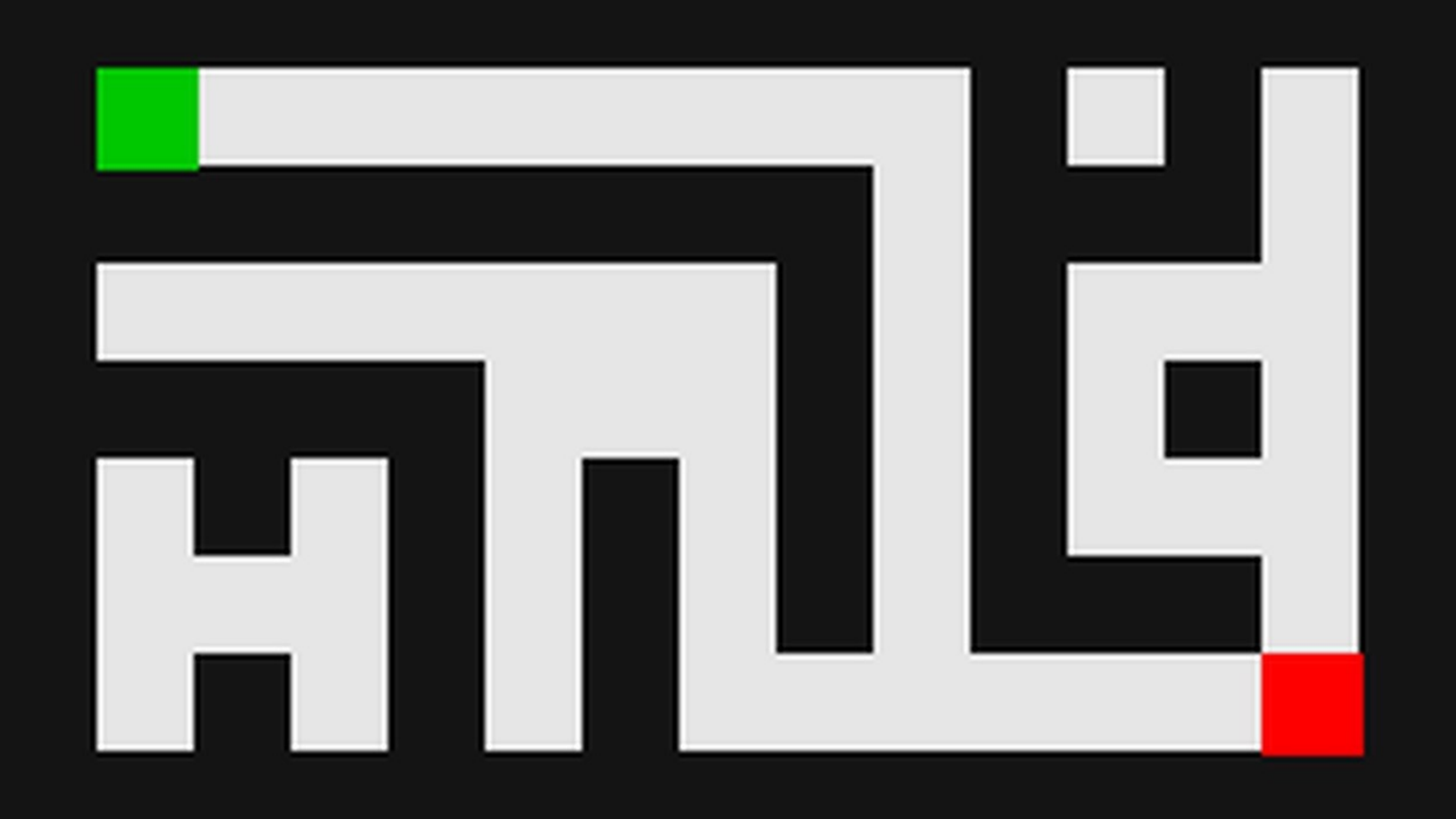}
{A top-down 2D maze. The black lines represent walls. the green square marks the starting point, and the red square marks the ending point. Draw a solid red line animates smoothly from the green square navigating the correct path through the maze structure without intersecting the walls, finally reaching the red square. Static camera, no pan, no zoom, no dolly, no perspective shifts.}

\subsection{Symbolic Reasoning}

\TaskEntry{Multimodal.}
{This task requires the model to solve a multimodal math problem presented on a whiteboard and generate a step-by-step written derivation leading to the final boxed answer. Images are sourced from existing multimodal understanding benchmarks like MMMU, MathVista and MathVision \citep{yue2024mmmu,lu2023mathvista,wang2024measuring}.}
{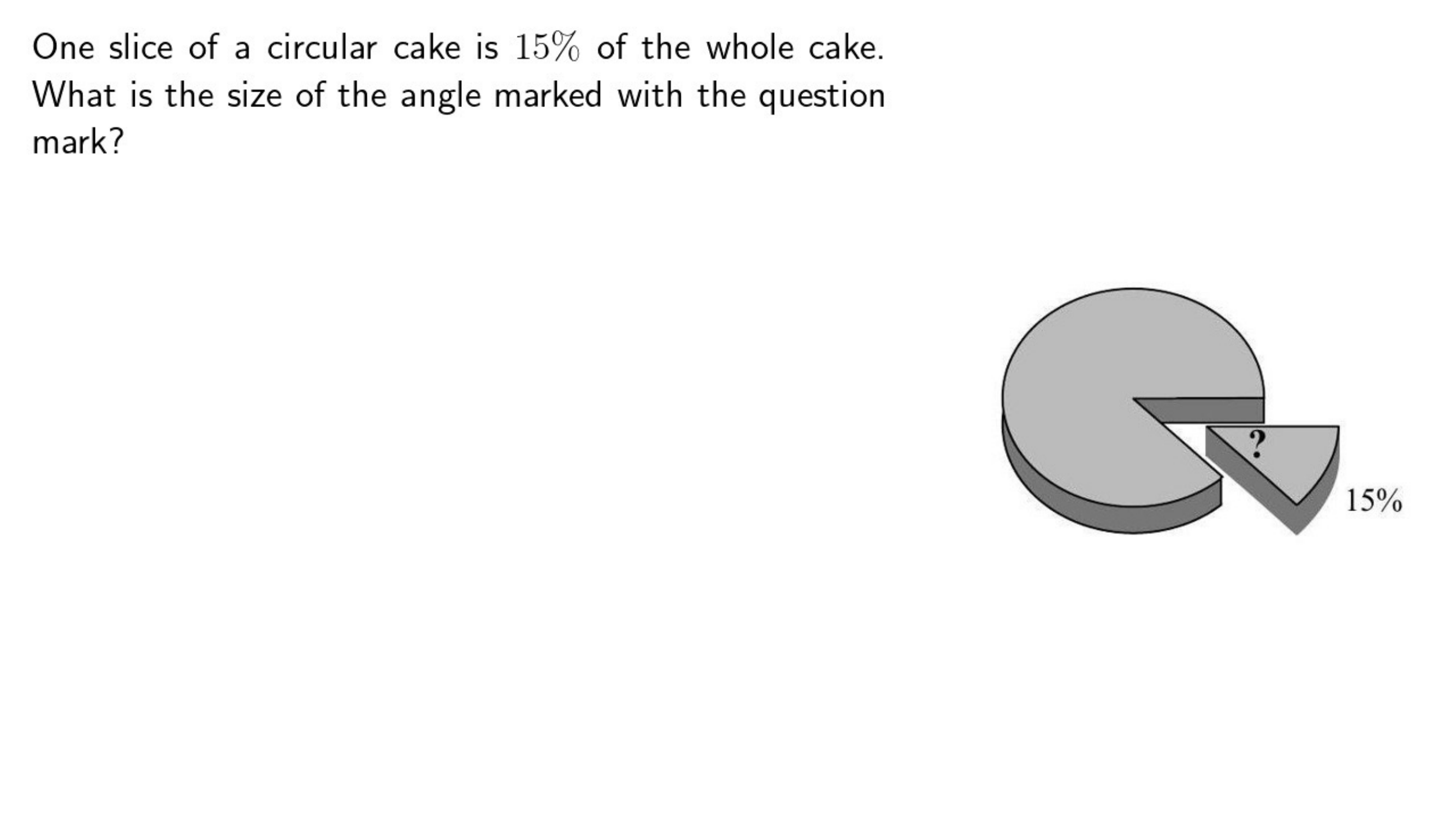}
{A static shot of a clean, well-lit whiteboard displaying the text or diagram for a multimodal problem: One slice of a circular cake is $15 \%$ of the whole cake. What is the size of the angle marked with the question mark?. The solution appears sequentially on the empty space below the problem, writing out the derivation process and concluding with the final answer clearly boxed. Static camera, no pan, no zoom, no perspective shifts, consistent scale.}

\TaskEntry{Math.}
{This task requires the model to solve a text-based math word problem shown on a whiteboard and generate a step-by-step written solution ending with a boxed final answer. Images are generated programmatically based on existing text-based math benchmarks like MATH-500 and GSM8K\citep{cobbe2021training, hendrycks2021measuring}.}
{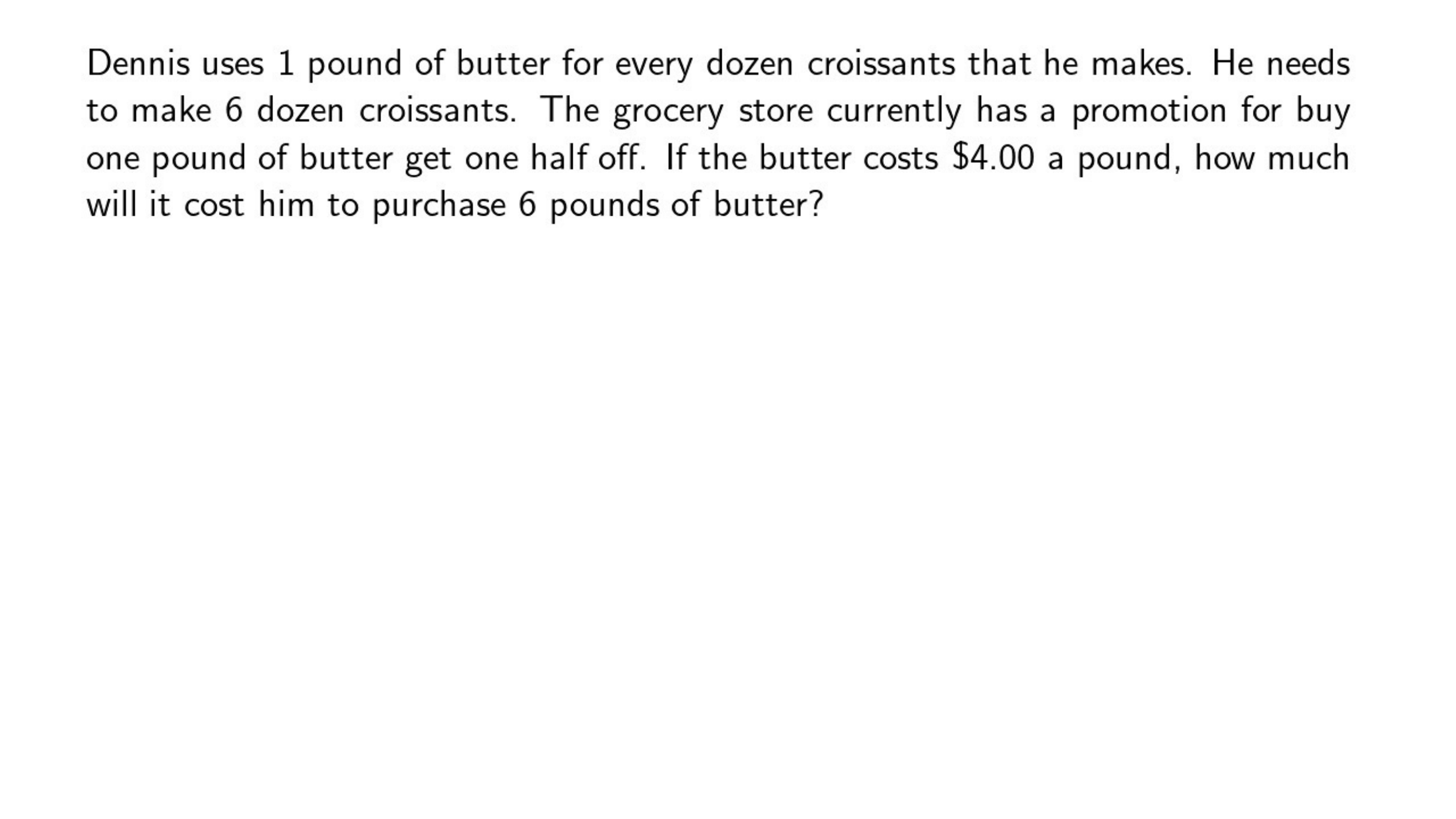}
{A static shot of a clean, well-lit whiteboard displaying the text or diagram for a math problem: Dennis uses 1 pound of butter for every dozen croissants that he makes.  He needs to make 6 dozen croissants.  The grocery store currently has a promotion for buy one pound of butter get one half off.  If the butter costs \$4.00 a pound, how much will it cost him to purchase 6 pounds of butter?. The solution appears sequentially on the empty space below the problem, writing out the derivation process and concluding with the final answer clearly boxed. Static camera, no pan, no zoom, no perspective shifts, consistent scale.}

\TaskEntry{Knowledge.}
{This task requires the model to answer a text-centric knowledge multiple-choice question shown on a whiteboard and write a step-by-step explanation culminating in a boxed final choice. Images are generated programmatically based on existing knowledge benchmarks like MMLU\citep{wang2024mmlu} and GPQA~\cite{rein2024gpqa}.}
{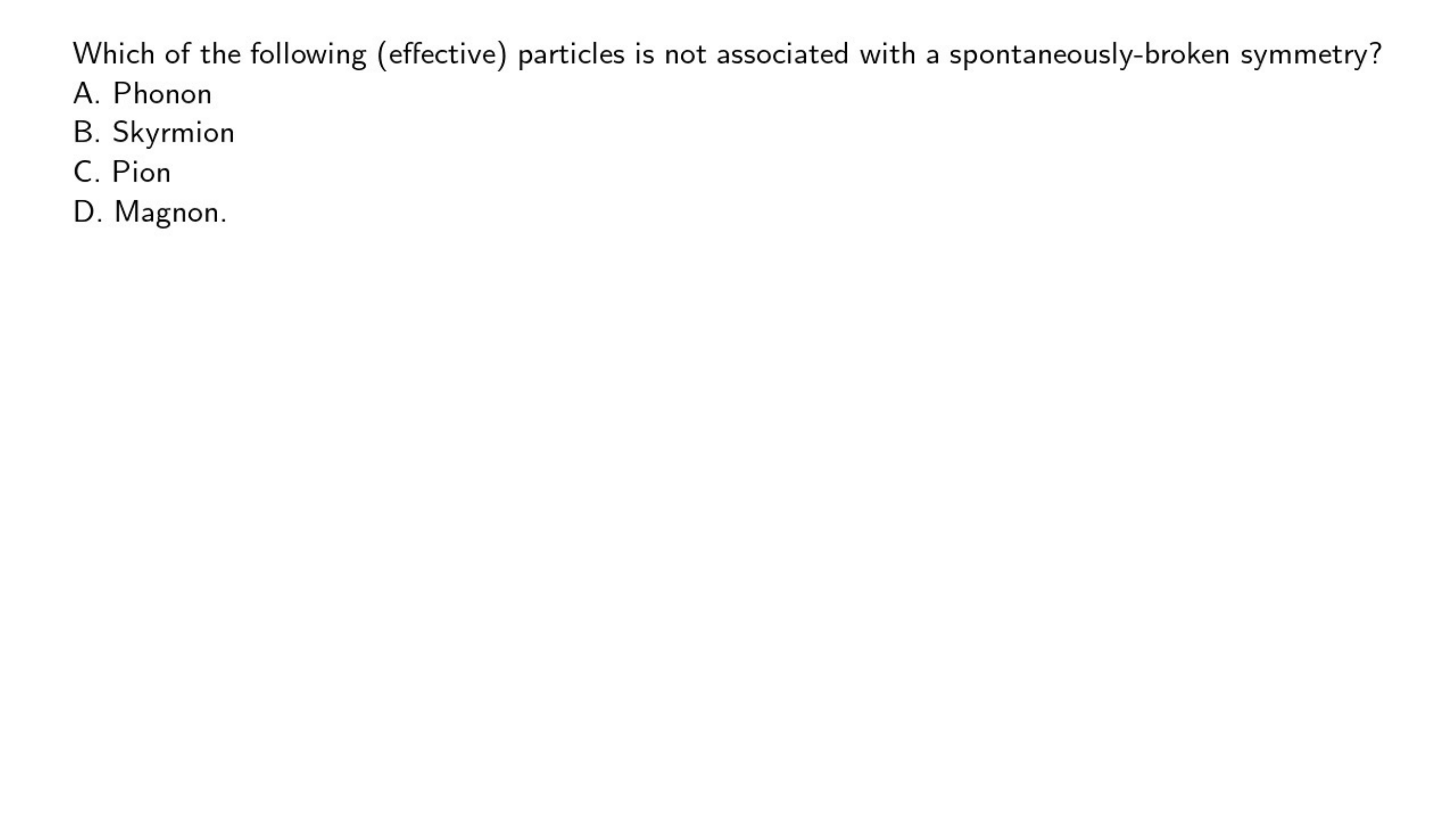}
{A static shot of a clean, well-lit whiteboard displaying the text or diagram for a knowledge problem: Which of the following (effective) particles is not associated with a spontaneously-broken symmetry?\\A. Phonon\\B. Skyrmion\\C. Pion\\D. Magnon.. The solution appears sequentially on the empty space below the problem, writing out the derivation process and concluding with the final answer clearly boxed. Static camera, no pan, no zoom, no perspective shifts, consistent scale.}

\subsection{Spatial Reasoning}

\TaskEntry{Block Rotate.}
{This task requires the model to render a multi-cube block structure while the viewpoint rotates horizontally by 180° around it, preserving correct spatial relationships between cubes. Images are sourced from VMEvalkit \citep{deng2025video}.}
{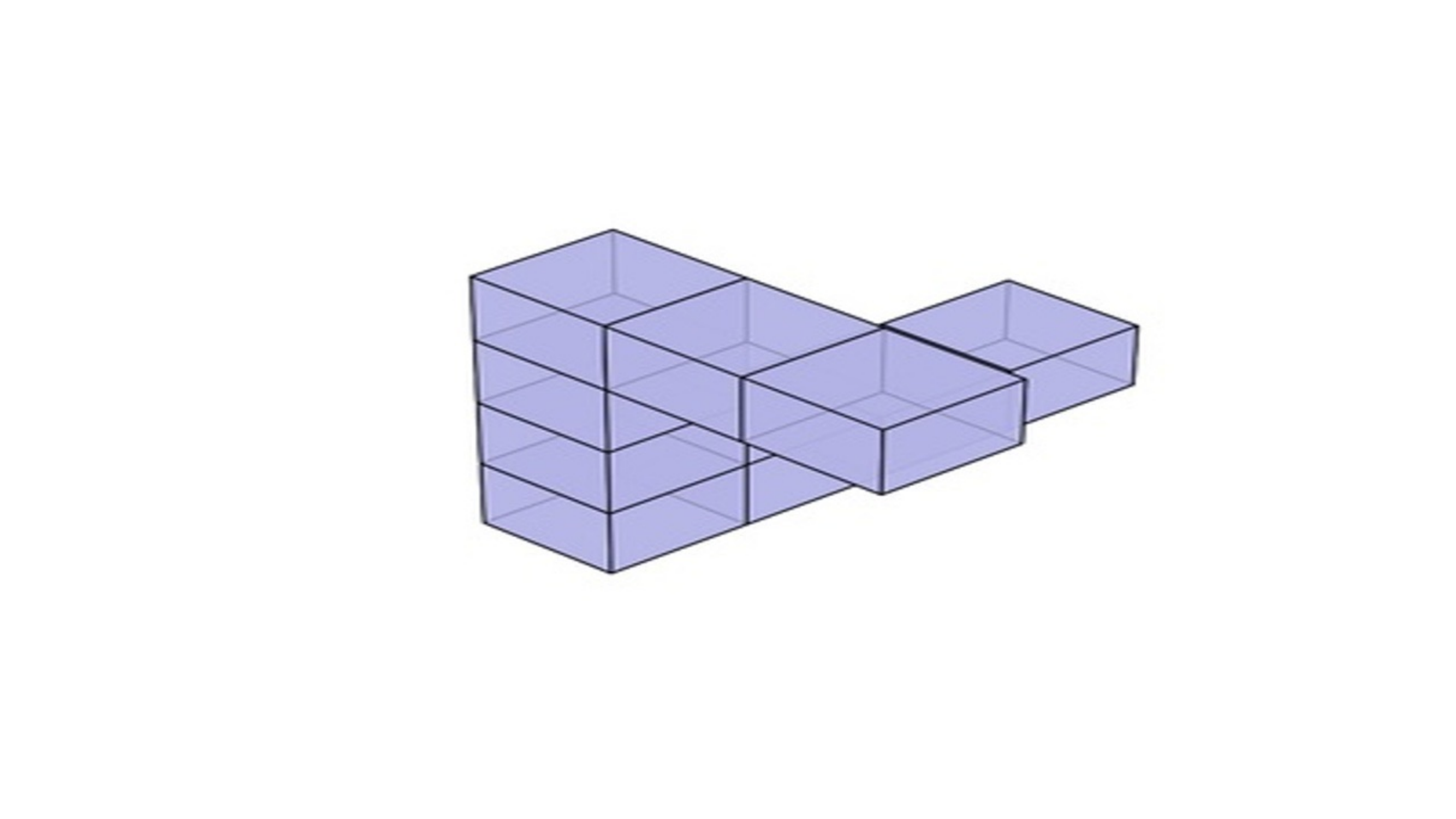}
{A structure composed of 9 cubes viewed from 316° azimuth, 40° elevation. The camera executes a smooth horizontal orbit around the structure, transitioning to a 136° azimuth which is a 180-degree rotation while maintaining a constant distance from the object. Smooth orbital movement, constant radius, no zoom, no vertical tilt, consistent scale.}

\TaskEntry{Dice Rolling.}
{This task requires the model to generate a video of a six-sided die rolling along a specified grid path via successive 90° edge flips, maintaining correct face orientation throughout. Images are rendered using Blender.\footnote{https://www.blender.org/}}
{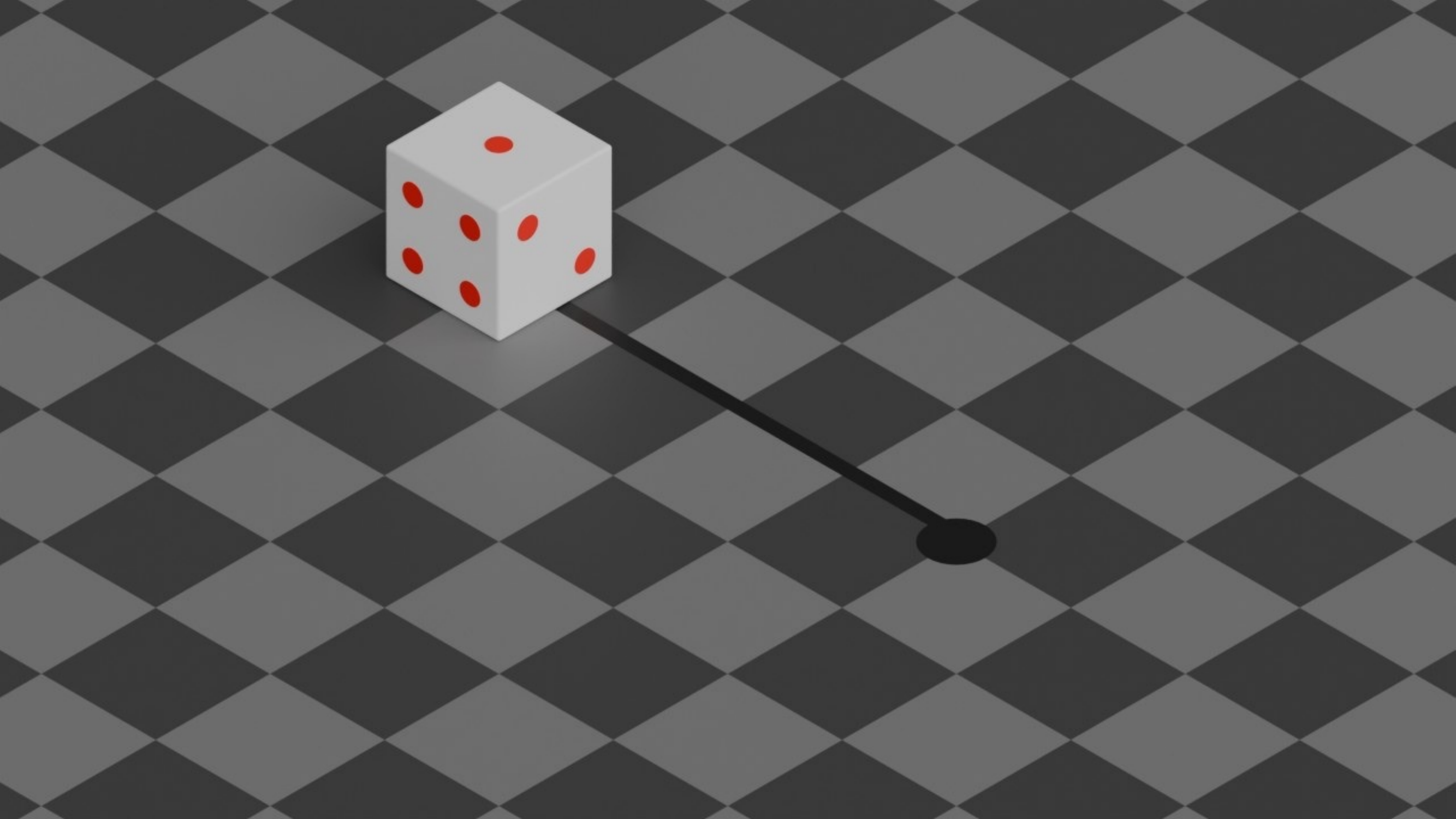}
{A standard six-sided die (faces arranged 1 opposite 6, 2 opposite 5, 3 opposite 4) resting on a grid-marked surface. A thick black path follows the grid lines, originating from the die's position and terminating at a solid black circular target. The die travels along the black path by executing precise 90-degree flips, pivoting on the edge in contact with the table for each step. The motion continues sequentially along the line until the die lands perfectly covering the black circular target. Static camera, high-angle view, sharp focus, no zoom, no pan, no perspective shifts.}

\TaskEntry{Image Restore.}
{This task requires the model to restore a shuffled image by translating sub-tiles into their corresponding dashed grid slots to reconstruct the full picture. Images are collected from open-source web sources.}
{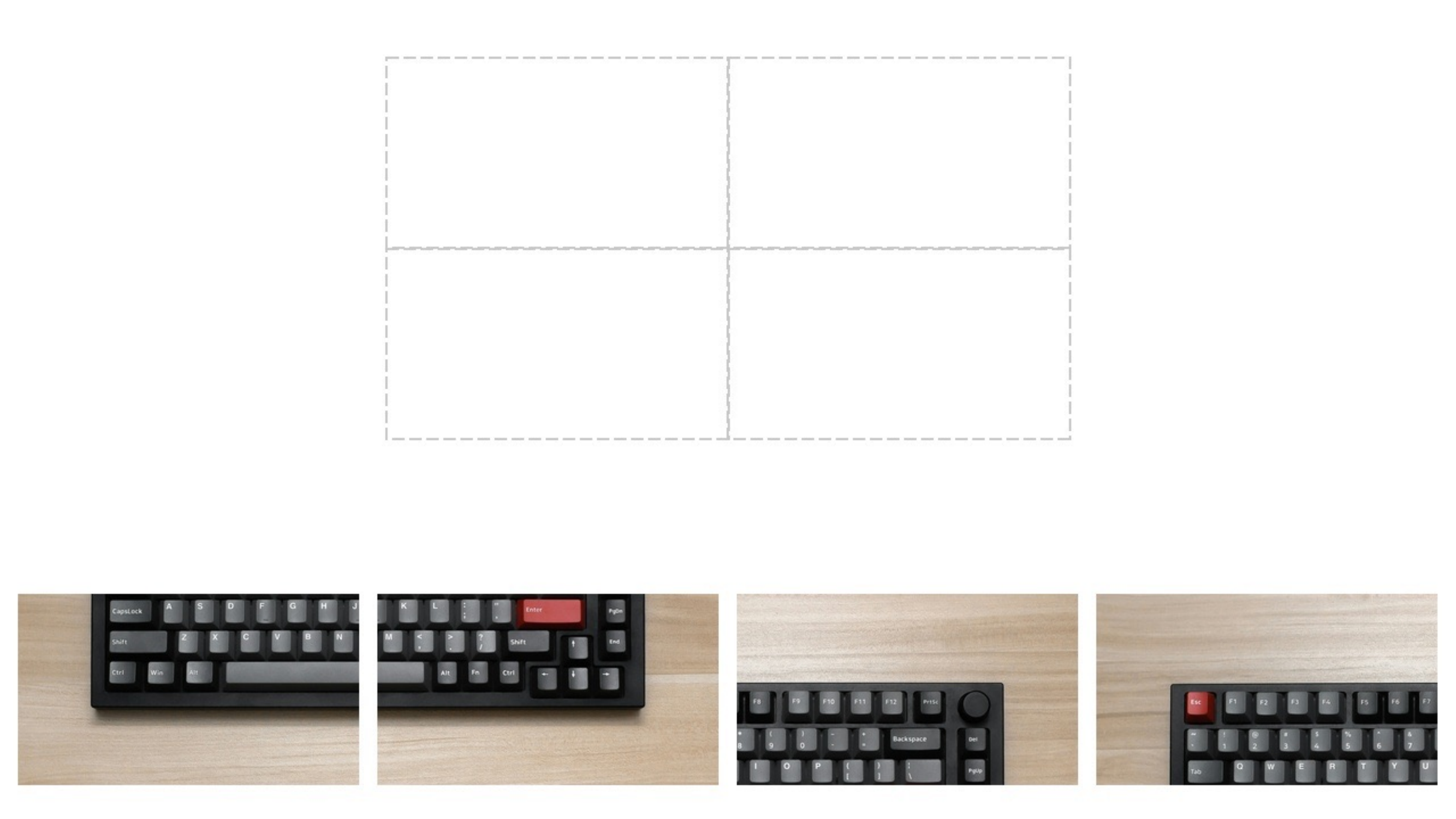}
{A clean white background. A set of shuffled image sub-tiles are arranged at the bottom of the frame, situated below a centered target grid of dashed outlines. The tiles translate smoothly from their starting positions into the matching dashed frames to fully reconstruct the original image. Flat 2D style, locked camera, no zoom, no pan, no perspective shifts.}

\subsection{Physics Reasoning}

\TaskEntry{Experiment.}
{Experiment task expects the model to render the process of a serires of classic physical experiment. The task test the ability of correct rendering of real world physical laws. The input image is maually collected from advanced image generation models.}
{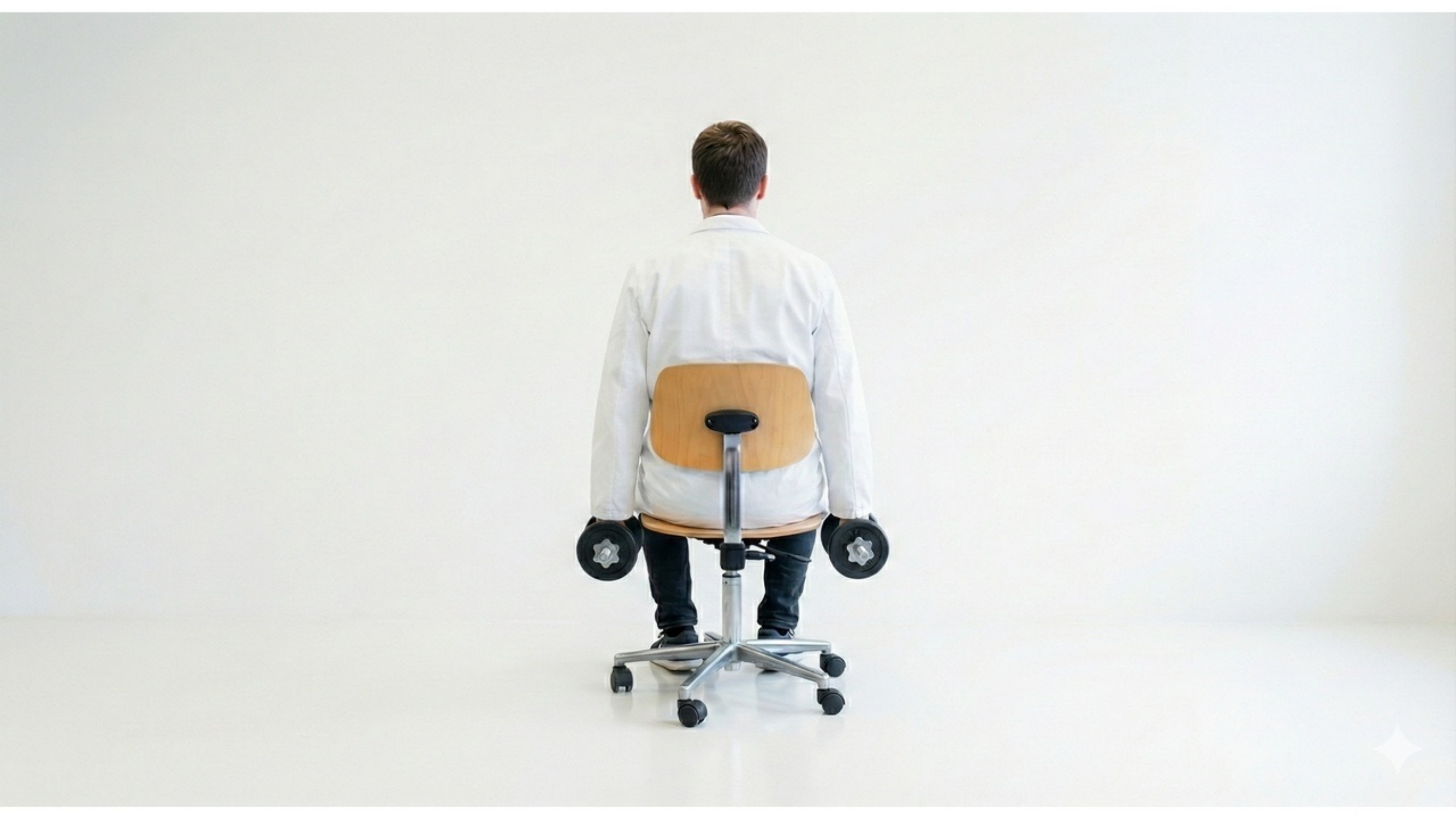}
{A person sits on a chair and keeps spinning. Initially, the arms are rest by his sides, then gradually rise and extend horizontally. Static camera, no pan, no zoom, no dolly.}

\TaskEntry{Game.}
{This task requires the model to simulate a mobile game scene where cutting specified ropes causes a suspended candy to move under game physics toward a character. Images are manually collected from existing mobile games.\footnote{https://en.wikipedia.org/wiki/Cut\_the\_Rope\_video\_game}}
{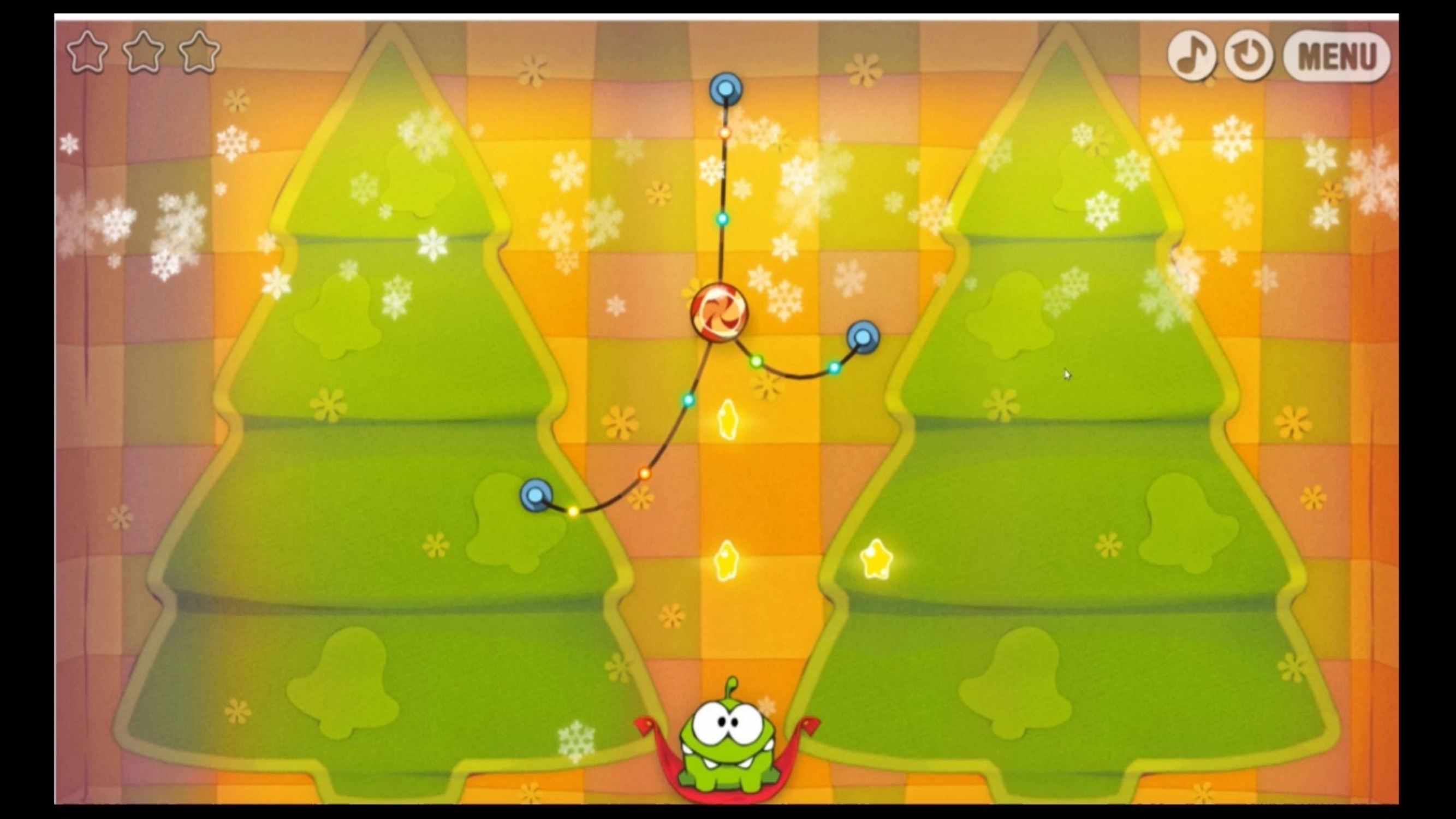}
{A mobile video game where a candy is fixed by several ropes, each rope connecting one end to the candy and the other end to the wall in the background. A cartoon character is at the bottom of the screen. The candy will be eaten if it moves near the character. Cut the two ropes at the bottom. Static camera, no pan, no zoom, no dolly.}

\subsection{Planning Reasoning}

\TaskEntry{Object Manipulation.}
{This task requires the model to render an object manipulation sequence where a robot arm grasps a target item and relocates it to a specified position on the table. Images are collected from RoboBench \citep{luo2025robobench}.}
{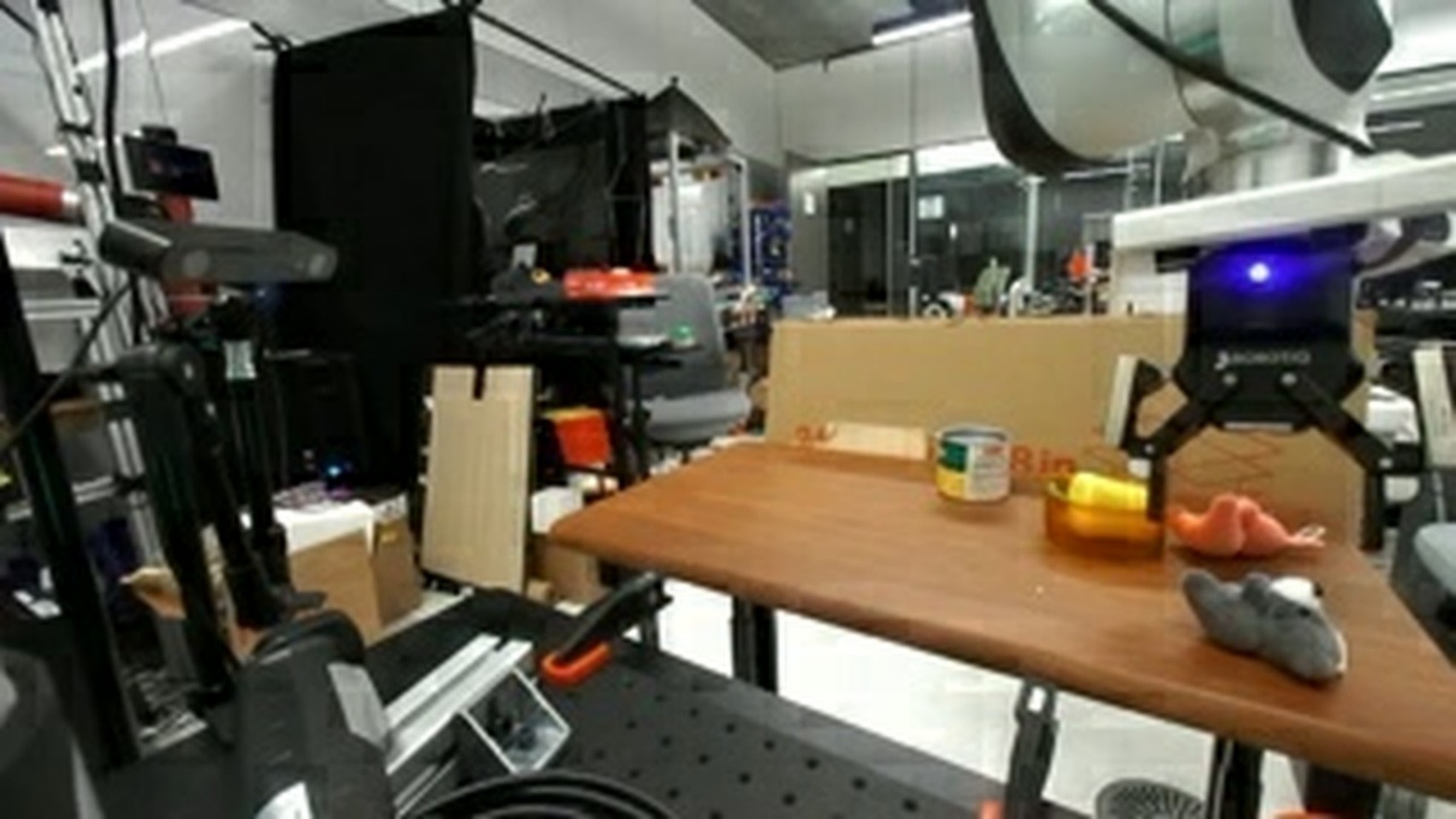}
{A lab workbench scene with a wooden table in the center. On the right side of the table there is a gray doll/plush-like object, with other small items nearby. A robot arm is positioned to the right of the table. The robot arm grasps the gray doll and places it down on the left side of the table. Static camera, no pan, no zoom, no dolly.}

\TaskEntry{Navigation.}
{This task requires the model to draw a continuous path on a top-down 2D map, navigating from a start location through specified waypoints to the destination. Images are collected from an open-source dataset.}
{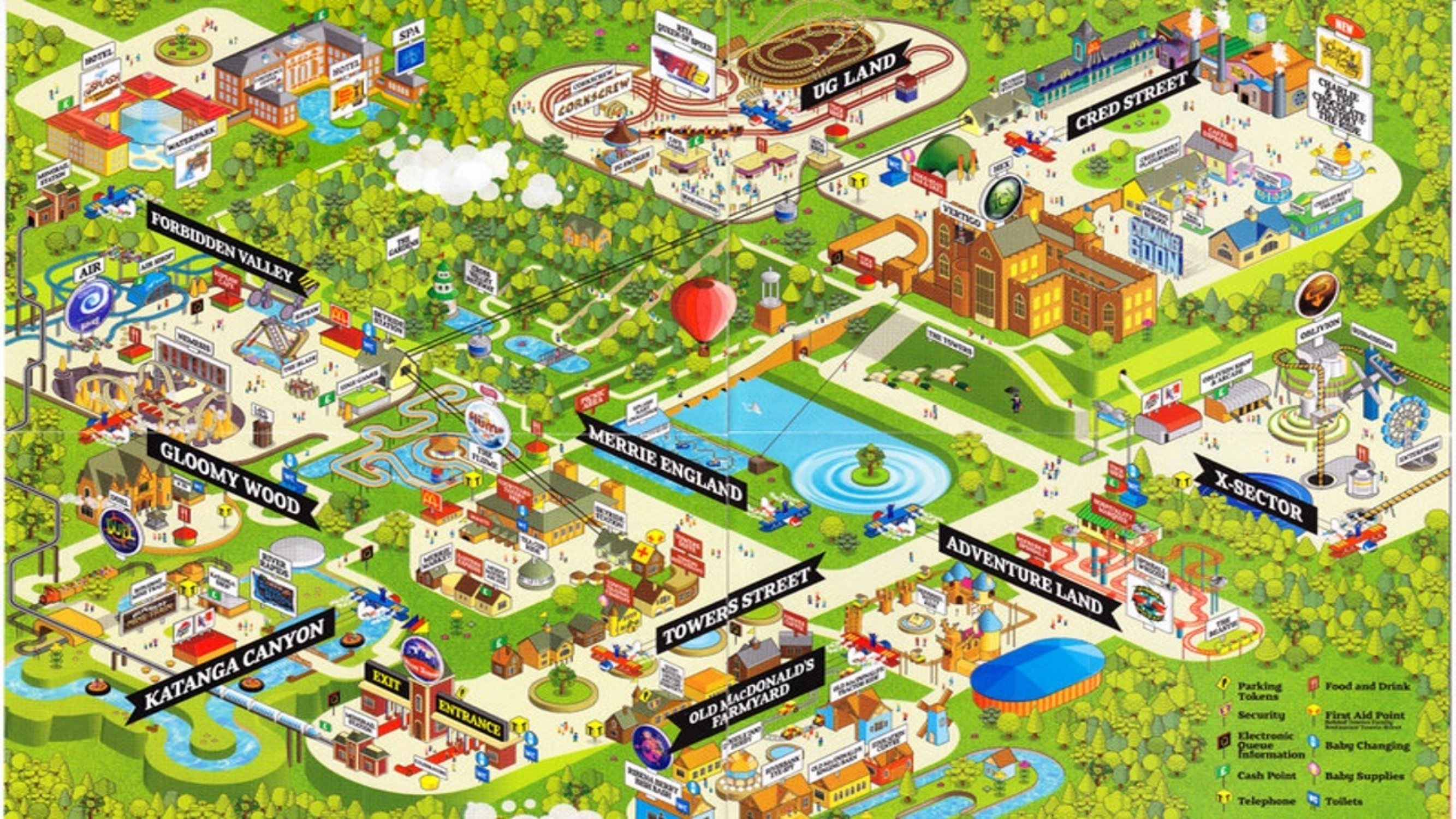}
{A 2D map with a top-down view. Draw a red path on the map to navigate from Air, passing through Corkscrew  and finally reaching Oblivion. The red path should be continuously drawn as a smooth line. Static camera, no pan, no zoom, no doll.}

\section{Rubric Examples}~\label{app:rubric}
In this section, we demonstrate the system prompt, all domain prompts and an example of task constraints.

\subsection{System Prompt}
\begin{center}
\begin{tcolorbox}[colback=gray!5!white, colframe=gray!55!black, width=0.98\linewidth, breakable, title={{System prompt}}]
{
    You are a video evaluation judge. Your task is to assess a video generated by a model, based on an initial image and a task description. Carefully follow the instructions below to ensure precise evaluation.\\

    \textbf{Inputs:}
    \begin{itemize}[left=0pt]
        \item \textbf{Domain Overview}: A brief introduction of the task's domain and evaluation focus.
        \item \textbf{Task Prompt}: The prompt provided to the video generation model.
        \item \textbf{Initial Image}: The image given to the model as the starting point for video generation.
        \item \textbf{Generated Video}: The video generated by the model based on the initial image and task prompt.
        \item \textbf{Process Constraints}: The set of process-level constraints that the generated video content should adhere to.
        \item \textbf{References}: Ground-truth information to assist in evaluation, if available. Could be videos, images, or textual descriptions.\\
    \end{itemize}

    \textbf{Evaluation Criteria:}\\
    Your evaluation should be based on the following two aspects:\\
    1. \textbf{Process Consistency}: Verify if \textbf{all frames} in the generated video adhere to the \textbf{process constraints}. Refer to the \textbf{References} if necessary.\\
    2. \textbf{Goal Consistency}: Verify if \textbf{at least one frame} in the generated video fulfills the \textbf{task prompt}. Refer to the \textbf{References} if necessary.

    The video is considered \textbf{correct} only if it fulfills both process and goal consistencies. If either is not fulfilled, the video is \textbf{incorrect}.\\

    \textbf{Output Format:}
    \begin{itemize}[left=0pt]
        \item Begin with your analysis of both process and goal consistencies inside \texttt{<think>} tags.
        \item After completing your analysis, provide your final evaluation inside \texttt{<answer>} tags.
    \end{itemize}

    Example:
    \noindent
    \begin{verbatim}
<think>
YOUR ANALYSIS HERE
</think>
<answer>
{
    "process_consistency":
    "goal_consistency":
    "decision": 
}
</answer>
    \end{verbatim}

    Please ensure your response follows this structure precisely.
}
\end{tcolorbox}
\end{center}

\subsection{Domain Introduction}
\begin{center}
\begin{tcolorbox}[colback=gray!5!white, colframe=gray!55!black, width=0.98\linewidth, breakable, title={{Domain: Temporal Reasoning}}]
{
    This task belongs to the Temporal Reasoning domain. The evaluation should focus on the model's ability to accurately track and represent the dynamic evolution of objects, scenes, and actions over time.
}
\end{tcolorbox}
\end{center}

\begin{center}
\begin{tcolorbox}[colback=gray!5!white, colframe=gray!55!black, width=0.98\linewidth, breakable, title={{Domain: Structural Reasoning}}]
{
    This task belongs to the Structural Reasoning domain. The evaluation should focus on the model's ability to process and generate content based on strict, predefined rules, constraints, and structured environments (e.g., games, puzzles, geometric layouts).
}
\end{tcolorbox}
\end{center}

\begin{center}
\begin{tcolorbox}[colback=gray!5!white, colframe=gray!55!black, width=0.98\linewidth, breakable, title={{Domain: Symbolic Reasoning}}]
{
    This video belongs to the Symbolic Reasoning domain. The evaluation should focus on the model's ability to interpret, manipulate, and generate abstract symbols (numbers, text, logical operators) to solve problems.
}
\end{tcolorbox}
\end{center}

\begin{center}
\begin{tcolorbox}[colback=gray!5!white, colframe=gray!55!black, width=0.98\linewidth, breakable, title={{Domain: Spatial Reasoning}}]
{
    This task belongs to the Spatial Reasoning domain. The evaluation should focus on the model's understanding of 3D space, geometric structures, perspective, and the relative positions of objects.
}
\end{tcolorbox}
\end{center}

\begin{center}
\begin{tcolorbox}[colback=gray!5!white, colframe=gray!55!black, width=0.98\linewidth, breakable, title={{Domain: Physics Reasoning}}]
{
    This task belongs to the Physical Reasoning domain. The evaluation should focus on the model's ability to simulate accurate physical laws, (e.g., gravity, collision dynamics, material properties).
}
\end{tcolorbox}
\end{center}

\begin{center}
\begin{tcolorbox}[colback=gray!5!white, colframe=gray!55!black, width=0.98\linewidth, breakable, title={{Domain: Planning Reasoning}}]
{
    This task belongs to the Planning Reasoning domain. The evaluation should focus on the model's ability to execute complex, multi-step tasks that require foresight, strategy, and logical sequencing.
}
\end{tcolorbox}
\end{center}